\documentclass{article}
\usepackage{arxiv}

\usepackage[utf8]{inputenc} % allow utf-8 input
\usepackage[T1]{fontenc}    % use 8-bit T1 fonts
\usepackage{hyperref}       % hyperlinks
\usepackage{url}            % simple URL typesetting
\usepackage{booktabs}       % professional-quality tables
\usepackage{amsfonts}       % blackboard math symbols
\usepackage{nicefrac}       % compact symbols for 1/2, etc.
\usepackage{microtype}      % microtypography
\usepackage{cleveref}       % smart cross-referencing
\usepackage{lipsum}         % Can be removed after putting your text content
\usepackage{graphicx}
\usepackage{natbib}
\usepackage{doi}

\usepackage{subcaption}
\usepackage{scalerel}
\usepackage{siunitx}

\usepackage{setspace}
\title{
  Visuospatial navigation from the bottom-up: without vestibular integration, distance prediction, or maps
}

\date{\today}

\usepackage{authblk}

\author[1]{
	{Patrick Govoni\thanks{Corresponding Author E-mail: \texttt{pgovoni21@gmail.com} (PG)}}
}
\author[1,2,3]{
	{Pawel Romanczuk}
}
\affil[1]{Institute for Theoretical Biology, Department of Biology, Humboldt Universität zu Berlin, Berlin, Germany}
\affil[2]{Science of Intelligence, Research Cluster of Excellence, Berlin, Germany}
\affil[3]{Bernstein Center for Computational Neuroscience, Berlin, Germany}

\renewcommand{\shorttitle}{}

\hypersetup{
pdftitle={Visuospatial navigation from the bottom-up: without vestibular integration, distance prediction, or maps},
pdfsubject={},
pdfauthor={Patrick Govoni, Pawel Romanczuk},
pdfkeywords={},
}

\begin{document}
\maketitle

\begin{abstract}

% BACKGROUND information: a sentence giving a broad introduction to the field comprehensible to the general reader
Navigation is believed to be controlled by at least two partially dissociable systems in the brain.
% then a sentence of more detailed background specific to your study
The cognitive map informs an organism of its location and bearing, updated by integrating vestibular self-motion or predicting distances to landmarks.
Route-based navigation, on the other hand, directly evaluate sequential movement decisions from immediate percepts.
% This should be followed by an explanation of the OBJECTIVES/METHODS
Here we demonstrate the sufficiency of visual route-based decision-making in a classic open field navigation task often assumed to require a cognitive map.
% then the RESULTS
Three distinct strategies emerge to robustly navigate to a hidden goal, 
each conferring contextual tradeoffs analyzed at both neural and behavioral scales,
as well as qualitatively aligning with behavior observed across the biological spectrum.
% final sentence should outline the main CONCLUSIONS of the study, in terms that will be comprehensible to all readers
We propose reframing navigation from the bottom-up, 
through an egocentric episodic perspective
without assuming online access to computationally expensive top-down representations,
to better explain behavior under energetic or attentional constraints.

% 125 words or less
% --> 123 words currently

\end{abstract}
% \end{frontmatter}


% \section{Author summary}

% Reliable navigation is crucial for mobile organisms. 
% Navigation is often believed to rely on cognitive maps — internal representations of space, driven by predictive or integrative means, that allow for planning and shortcuts. 
% However, we show that simple, response-based strategies, driven by immediate visual cues, are sufficient for robust navigation in open fields. 
% Using a computational model, we trained agents to navigate to a hidden goal using minimal visual information. 
% Surprisingly, three distinct navigation strategies emerged, each with unique tradeoffs in efficiency and robustness. 
% These strategies parallel behaviors observed across the biological spectrum, 
% suggesting that response-based strategies may be fundamental to navigation. 
% Our findings challenge the assumption that complex cognitive maps or even distance calculation are necessary for navigation 
% and highlight the importance of taking a bottom-up approach in understanding how organisms behave. 

\section{Introduction}

% % foraging : search + navigation : environmental predictability
% As dissipative systems, all biological beings are bound by the curse of having to find food. 
% While some have evolved sedentary strategies, others need to move to ensure a steady flow of nutrients.
% When food location is unpredictable, this movement can be classified as a search problem 
% \cite{haydenNeuronalBasisSequential2011, wispinskiAdaptivePatchForaging2023}. % benichou, davidson
% Organisms otherwise living in environments with regularity in food location can benefit from the ability to return to known resource-abundant areas, 
% e.g. a bee to a flower patch, a mouse to the kitchens of an apartment building, or a human foraging for mushrooms.
% Simply put, the organism must learn to navigate. 

% navigation + integ/pred
Navigation involves correlating perceivable spatial features with expected goal location, 
whether it be food, shelter, or other area of interest,
requiring the environment to be sensed and processed before effective movement. 
% maps
Cognitive maps, widely accepted as the ultimate way many organisms represent space, 
describe a system of hippocampal cells tuned to distinct spatial elements collectively composing a Euclidean graph
upon which direct pathways and novel shortcuts can be calculated via simple vector transformation
\cite{epsteinCognitiveMapHumans2017, bellmundNavigatingCognitionSpatial2018a}. % behrens
Although maps can be used for useful top-down planning, particularly regarding shortcuts, 
focus on this subset of navigational possibility has left its alternatives relatively understudied.
% routes
Route-based navigation, on the other hand, concatenates sequences of egocentric movement decisions based on local observations,
which can offer greater efficiency in complex environments 
\cite{peerStructuringKnowledgeCognitive2021, parra-barreroMapSpatialNavigation2023},
or afford low demand reliability allowing energy to be directed to more critical or less predictable tasks
\cite{prietoPlanningAutopilotAssociative2023}. % harms
Following familiar routes may rely on the bottom-up application of episodic memory
\cite{igloiLateralizedHumanHippocampal2010, goodroeComplexNatureHippocampalStriatal2018, 
wangEgocentricAllocentricRepresentations2020, kunzNeuralCodeEgocentric2021, 
mooreLinkingHippocampalMultiplexed2021, parra-barreroMapSpatialNavigation2023},
a second key hippocampal function,
while weaving together a network of routes enable construction of a non-Euclidean topology via state transitions
\cite{warrenNonEuclideanNavigation2019, lanGoaldirectedNavigationHumans2025}.

% separability bw map-based vestibular/distance + vision
Map-based navigation algorithms often require online distance prediction
\cite{recanatesiPredictiveLearningNetwork2021, kesslerHumanNavigationStrategies2024, stocklLocalPredictionlearningHighdimensional2024},
or otherwise vestibular integration
\cite{baninoVectorbasedNavigationUsing2018, cuevaEmergenceGridlikeRepresentations2018},
in order to function.
However, key visual dimensions were found to be functionally irrelevant for the two algorithms among those listed that performed ablation studies
\cite{baninoVectorbasedNavigationUsing2018, recanatesiPredictiveLearningNetwork2021},
suggesting that vision provides a redundant information stream.

% vision > vestibular
In contrast, many animals have been shown to principally rely on visual cues for reliable navigation.
Gerbils and hamsters tend to use configurations of visible landmarks when well-lit, conspicuous, and reliable, 
otherwise resorting to their sense of self-motion \cite{collettLandmarkLearningVisuospatial1986, jefferyNeurobiologySpatialBehaviour2003}.
Similarly, the primate tendency to fixate our eyes rather than heads during self-motion 
may prioritize visual conflict detection over vestibular integration
\cite{SlownessSparsenessLead, garzorzVisualVestibularConflictDetection2017},
which would explain stronger hippocampal cell correlations to head orientation and visual perspective than to spatial position
\cite{rollsSpatialViewCells1997, ekstromCellularNetworksUnderlying2003, jacobsSenseDirectionHuman2010, maoSpatialModulationHippocampal2021, pizaPrimacyVisionShapes2024}.
Even rats, with normally strong positional hippocampal correlations, 
lose spatial selectivity in virtual reality
despite maintaining directionality~\cite{aghajanImpairedSpatialSelectivity2015, acharyaCausalInfluenceVisual2016}
and flexible navigation~\cite{mooreLinkingHippocampalMultiplexed2021}.
That is, restricting the vestibular sense forces a sequential route-based strategy grounded in vision. 

% simple vision > distance-based vision
Although visual distance calculation may be highly useful in specific contexts,
e.g. cliff avoidance, prey capture, or jumping gaps~\cite{saleemInteractionsRodentVisual2023},
whether in the natural environment or in designed experimental paradigms,
this perceptual dimension may not be as universal as assumed, or as accessible without considering its cost.
Eliminating depth does not significantly affect visual orientation capability in hamsters~\cite{etienneNavigationSmallMammal1992}.
Rats prioritize overhead binocular fusion over constant frontal depth perception,
as the latter is not as critical for predator avoidance~\cite{wallaceRatsMaintainOverhead2013}.
Additionally, complex 3D habitats may be required to justify its development~\cite{hagbiHeightsPlainsHow2022}.

% personal/human context + feedforward suggestion
On our way to work or the market, routes we know quite well,
to what extent do we continually update distances to landmarks, or count our turns,
rather than rely on relative positions of cues in our visual field?
That allocentric map-based abilities both take longer to acquire and subside faster than egocentric navigation over the span of a lifetime
suggests the preeminence of route-based decision-making in the context of constrained cognitive capacity
\cite{colomboEgocentricAllocentricSpatial2017, tansanSpatialNavigationChildhood2022}.
Rather than assuming unabated access to high dimensional representations and abilities, 
perhaps a simpler feedforward framework can explain a subset of navigational behaviors across the biological spectrum,
which, given nature's propensity to minimize energy, may be favored if performance is adequately robust.

% approach + MWM ambiguity + minimal:generalizable
In pursuit of this hypothesis,
minimal visual perception-action loops of embodied agents, without predictive or integrative feedback, 
are trained with fixed environment, perceptual, cognitive, and action constraints
to solve a classic spatial navigation task.
Although the Morris Water Maze with random initialization is often described as testing map- or place-based navigation strategy
involving vector calculation with respect to the hidden goal 
\cite{wolbersChallengesIdentifyingNeural2014, eichenbaumRoleHippocampusNavigation2017, 
vikbladhHippocampalContributionsModelBased2019, parra-barreroMapSpatialNavigation2023}, 
we intend to demonstrate how this task may be simply solved by egocentric visual observations and sequential routes.
Furthermore, by minimizing assumptions specific to certain organisms, our conceptual approach seeks to maximize generalizability
to discover fundamental properties of visual route-based navigation.

% 4602 chars, 3946 chars (nospaces), 655 words

\section{Results}

\subsection*{Model design}

% environment + task
The environment is designed to be minimally complex: a square grid with four distinct walls and a hidden circular foraging patch or goal. 
Using boundary walls restrains ability to estimate distance,
forcing reliance on relative angles to grid corners as salient landmarks.
Agents are initialized in random locations and orientations for each simulation.
As intermediate temporal representations are purposefully left out in our study,
they must instead rely on immediate visuospatial calculations to robustly reach the patch.

\begin{figure}[htb]
	\centering
	\includegraphics[width=.6\linewidth,trim={0 0 0 0},clip]{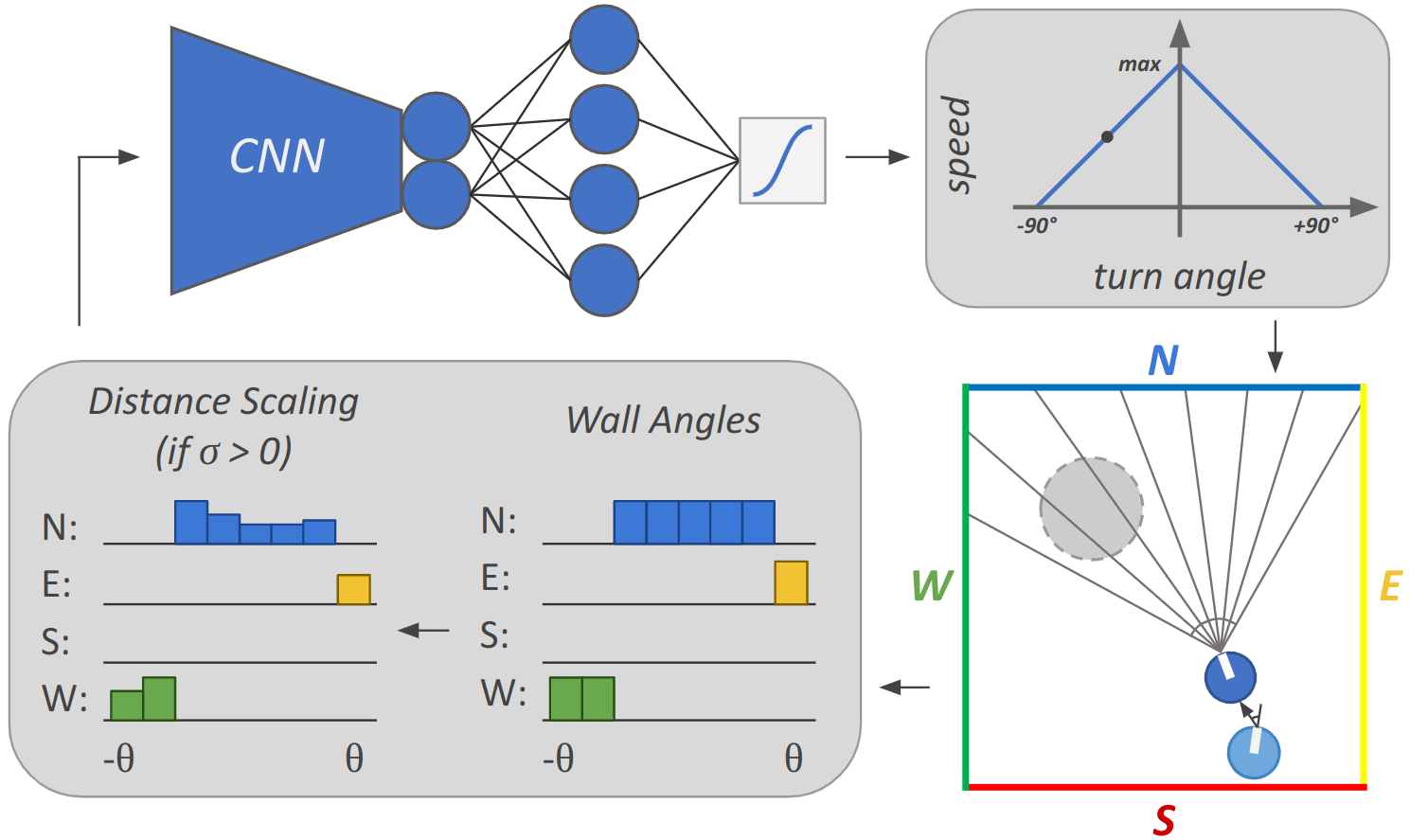} % LBRT crop
	\caption{
		\textbf{Agent perception-action loop flow.}
		Clockwise from bottom left: visual encoding, information processing, action conversion, environment update.
		Visual encoding consists of identifying walls corresponding to retinal angles of a raycast 
		($\upsilon$ number rays between -$\theta$ \& $\theta$ field of view limits) for minimal angle-only vision, 
		then adding distance information according to scaling factor $\sigma$.
		Visual information passes through convolutional neural network, perceptron, linear output layer, and hyperbolic tangent tranformations 
		to directly represent turning angle as well as speed via a linear function, 
		which updates agent position and orientation for the next timestep (single update shown in environment and black point on linear function).
		}
	\label{fig:flow}
  \end{figure}

% visual encoding
Inspired by~\cite{bastienModelCollectiveBehavior2020}, 
we simulated visual input as a raycast matrix, one-hot encoded per ray, extending from the agent to the boundary walls
(Fig~\ref{fig:flow}). 
Minimal vision provides information of identity and relative retinal angle of preceived walls, which we refer to as angle-based vision. 
For more complex vision, distance to the identified wall is logarithmically scaled into the encoding according to the Weber-Fechner law.
In an unknown environment, accurate distance estimation requires higher order computation,
e.g. stereopsis afforded by binocular fusion, static cues, dynamic temporal inference, or memory.
Angle-based vision, lacking distance information, represents minimal environmental information, 
constrained either from an estimation reliability or a cognitive load, attentional-based perspective.
Vision with distance information, scaled in at increasing signal variance ranges, characterizes increasing perceptual ability or cognitive bandwidth,
which may include predictive or integrative inferential capacities.

% architecture
The information processing flow of an agent consists of three modules:
a convolutional network (CNN), a single-layer perceptron, and a linear output layer.
A CNN is chosen mainly for its image processing ability, 
though in part to its conceptual similarity and quantified correlation to the visual cortex
\cite{lindsayConvolutionalNeuralNetworks2021}.
A simple single layer feedforward downstream network was found to confer sufficient performance,
whereas more complex architectures were found to provide minor additional benefit 
(\ref{fig:evol_params} Fig).
Finally, a single network output represents a ratio between turning angle and speed,
a minimal action space constraint describing the need to slow down in order to turn.
Note, this network is intended to be only phenomenologically related to sensorimotor information flow, 
rather than for specific biological detail.

% optimization
The network is optimized via an evolutionary strategies (ES) algorithm.
ES uses population-based black-box optimization without explicit gradient calculation, unlike reinforcement learning (RL) which optimizes a single agent
by using backpropagation on a continuous differentiable objective function
% , often via carefully engineered reward distributions and an intermediate value-based representation. 
% ES entails less complexity than RL, as there is no need for differentiability, value approximation, or within-trial credit assignment 
\cite{salimansEvolutionStrategiesScalable2017, majidDeepReinforcementLearning2021}.
% The main drawback of ES, sample inefficiency, is acceptable given its parallelizability.
Though both being mechanisms for optimization through perception-action-based interaction in the environment, 
they both result in the construction of useful learned representations 
\cite{anggrainiNeuralSignaturesReinforcement2018, bermudez-contrerasNeuroscienceSpatialNavigation2020, vijayabaskaranNavigationTaskAction2022}.

Fitness, or performance, was measured as the time taken to reach the patch plus the remaining distance if the patch was not reached by the end of the simulation. 
The first term is the primary driver for judging how well the agent navigates and the second guides initial learning behavior.
The relative weighting of each was varied in initial experiments but found to not make a significant difference.

\subsection*{Training}

% fitness: angle-only vs dist-embedded
Training the networks to this navigation task revealed the sufficiency of route-based navigation in an open field
(\ref{fig:evol_params} Fig).
Although only visual angle input is needed to adequately learn the task,
performance and speed of convergence improves relative to the salience of an added distance signal.
At maximal distance input ($\sigma = 1$), performance approaches the theoretical lower bound of perfect trajectories straight to the patch.
Attenuating its signal gradually reduces utility of the distance information, 
describing a titration between angle-based and distance-based perceptual encodings.
At $\sigma = 0.2$ signal variation is too subtle for the network to effectively use the information,
slowing down learning rate and hindering performance,
while at $\sigma = 0.1$ the signal is not evidently used at all.

\begin{figure}[htb]
	\centering
	\includegraphics[width=\linewidth,trim={0 0 0 0},clip]{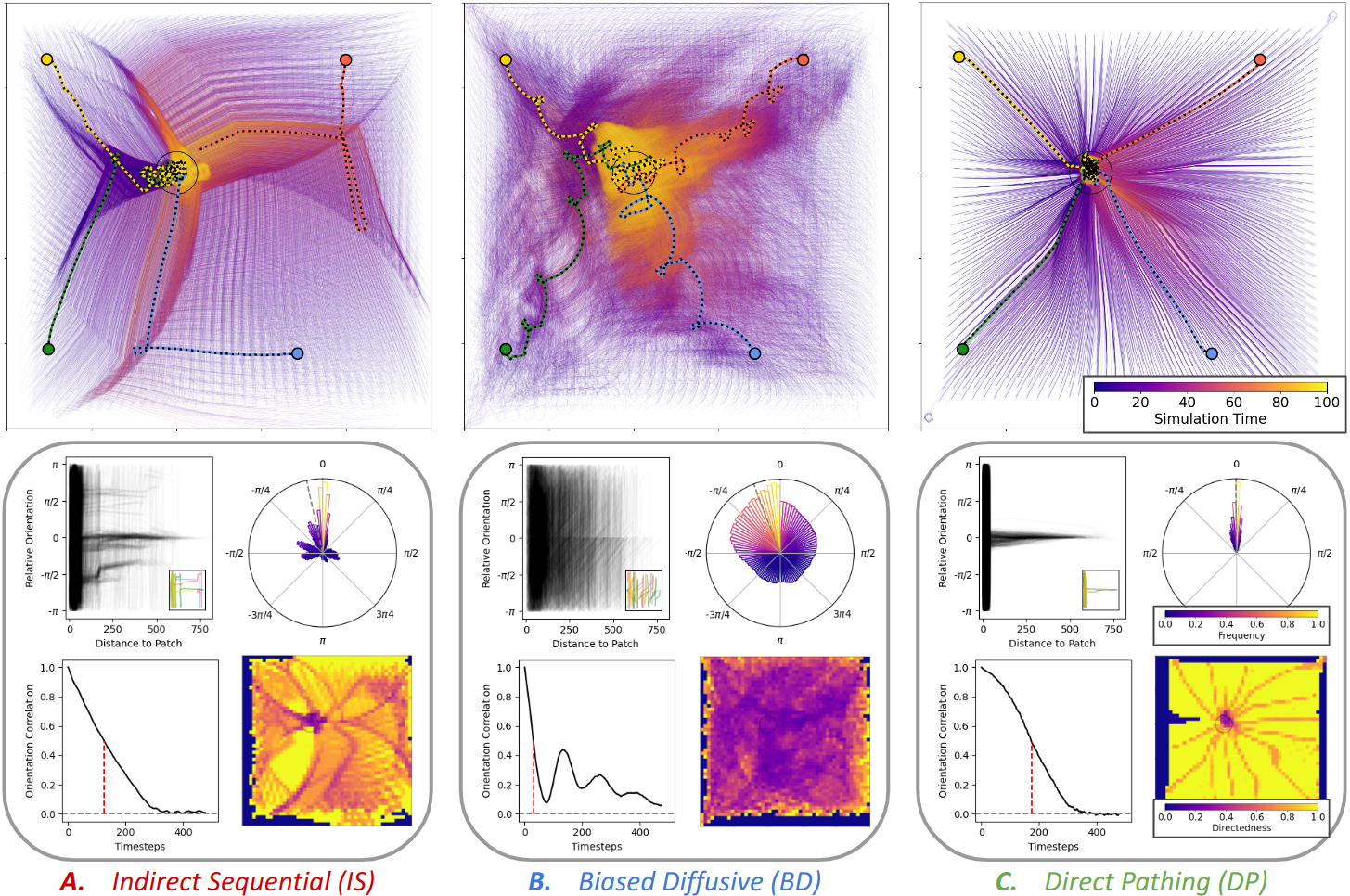}
	\caption{
	  \textbf{Three navigational classes, movement behavior \& correlations.}
	  \textit{Top row:} global movement behavior of three individual evolutionary runs or agents, 
	  with angle-based vision for A/B and with added distance scaling for C ($\sigma = 1$),
	  each with an 8 ray visual resolution ($\upsilon = 8$).
	  Solid lines: single agent trajectories from unique initial positions and orientations,
	  temporally colored, density and darkness reflect common routes.
	  Dotted lines: 4 example individual trajectories.
	  Black circle: patch location.
	  \textit{Bottom row:} movement correlations.
	  Top left: spatial heading profile with respect to initial heading and distance from patch center, inset shows same 4 example trajectories in above plots.
	  Top right: polar histogram of relative orientation for timesteps when agent is over 100 grid units away from patch center, with frequency proportional to area and color of bin.
	  Bottom left: temporal persistence of initial heading angle, marked by decorrelation time (50\% threshold, red dashed line), sinusoidal shape reflecting correlated oscillations.
	  Bottom right: directedness heatmap, an information theoretic measure calculated for each spatial bin in the environment.
	  See \ref{fig:trajs_ext} and \ref{fig:corrs_ext} Figs for other individuals.
		}
	\label{fig:trajs_corrs}
  \end{figure}

% intro to classes
On the behavioral side, a convergence into distinct classes was observed in the navigation trajectory space 
(Fig~\ref{fig:trajs_corrs} top row, movies~S1-S3). 
Two separate angle-only navigational strategies emerged despite identical constraints,
best described by the terms "indirect sequential" and "biased diffusive".
Adding distance scaling to the visual input resulted in a third class called "direct pathing".
Hybrid strategies exist, indicating a continuum among clusters
(\ref{fig:maps_IS_DP_BD} Fig), 
where exact boundaries depend on classification criteria details, discussed in Materials and methods.

\subsection*{Angle-only navigation: behavior \& mechanisms}

% IS behavior
The indirect sequential algorithm or class describes agents that learn to travel indirect trajectories to the patch, 
composed of a sequence of straight segments
(Fig~\ref{fig:trajs_corrs}a).
Elliptical arcs sweeping across the arena define transition points, scaffolding the route,
which the agent uses to shift direction through a single sharp turn.
Despite lacking distance perception, 
the agents learn to use visuospatial invariances afforded by the environmental structure in order to reach the patch.
However which invariances to use and how to compose them into an effective strategy varies between runs
(\ref{fig:trajs_ext} Fig).

\begin{figure}[htb]
	\centering
	\includegraphics[width=\linewidth,trim={0 0 0 0},clip]{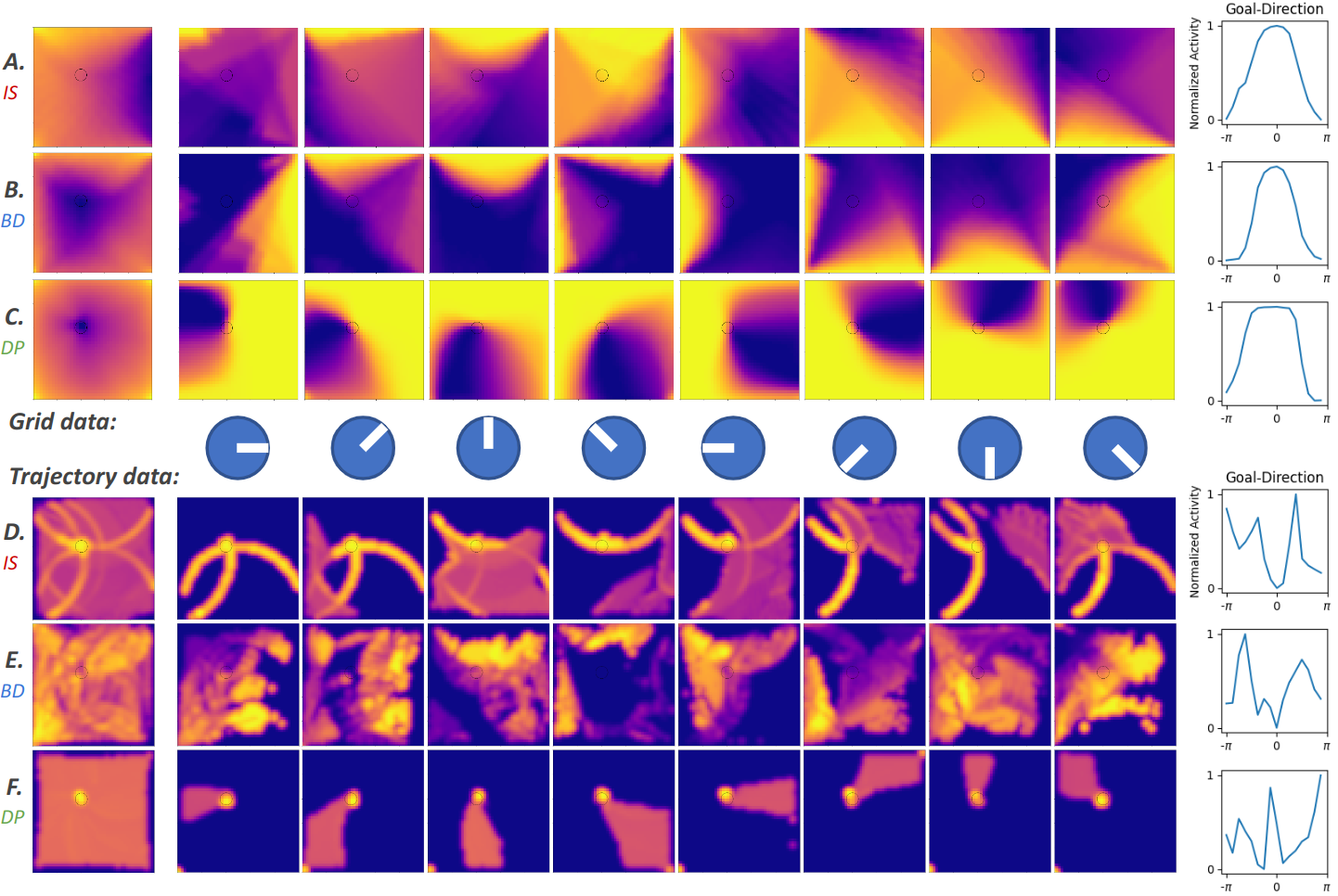}
	\caption{
	  \textbf{Distinct approaches to represent neural data.}
	  \textit{Top:} neural activity data gathered with uniform spatial and orientational occupancy.
	  \textit{Bottom:} agents initialized as above, data gathered over 500 timesteps and binned in an equivalent manner.
	  Left: spatial selectivity regardless of orientation, normalized from zero to maximum activation.
	%   2nd column: same as left, displaying mean orientation of each bin as arrow direction and mean vector length as color and size.
	  Middle eight: spatial selectivity with respect to orientation, illustrated by agent directions.
	  Right: neural tuning curves with respect to egocentric angle from the patch.
	  A-C \& D-F relate to the three agents and navigational classes of Fig~\ref{fig:trajs_corrs}:
	  indirect sequential (IS), biased diffusive (BD), and direct pathing (DP).
	}
	\label{fig:neureps}
  \end{figure}

% IS mechanism
The route-determining elliptical manifolds are governed by a variety of environmental, perceptual, and movement constraints, coordinated to the patch location
(\ref{fig:perturbs} and \ref{fig:map_rays} Figs, see Materials and methods).
Mechanistically, the agent learns to tie major turn decisions to specific views with respect to two adjacent corners.
As one of these views is rotated about the central axis of the respective wall, its decision threshold takes an elliptical form
(\ref{fig:map_rays} Fig, 1b).
While these manifolds cannot be easily seen by examining static neural activity normalized across location and orientation
(Fig~\ref{fig:neureps}A),
% as is common procedure in identifying neural correlations
% \cite{fdsa},
the ellipses are clearly evident when using goal-directed temporal data
(Fig~\ref{fig:neureps}D).
The goal location does not appear to spatially correlate with uniform grid data yet strongly overlaps with the visible manifolds in trajectory data
(Fig~\ref{fig:neureps}A \& D, left).
However, directional components are not evident when deriving neural tuning curves for the former while strong 90 degree components arise in the latter
(Fig~\ref{fig:neureps}A \& D, right).
% which suggests the indirect sequential strategy relies on subsampling the potential state space.
% More work is needed to fully understand the perception-action mechanism of how the agent may follow these routes
% and deemed to be out of scope for the current study.

% BD behavior
The movement profiles of the indirect sequential agents stand in contrast to the second class of algorithms, 
which can be distinguished by looping trajectories
(Fig~\ref{fig:trajs_corrs}b).
While overall progress is biased directly toward the patch, 
these agents regularly spin or turn, sensing and acting upon a wider set of observations.
The trajectory maps render clouds of agent trajectories with little apparent correlation among each other, 
revealing no distinct manifolds or sequences, instead the process is diffusive.

% BD mechanism
Curving trajectories dynamically sample the entire range of orientations, obscuring movement patterns.
Though its neural representation (Fig~\ref{fig:neureps}B)
and vector field of action output (\ref{fig:traj_vecfield_ori} Fig)
show how learned gradients, 
scaffolded by the same elliptical manifolds, 
drive overall movement toward the patch,
a mechanism that does not explain indirect sequential movement whose straight paths rely on a limited subsample of potential orientations.
Goal location does correlate with averaged grid data for these agents, while the trajectory data seems less useful to describe mechanism.
From a perturbative point of view, adding noise to the visual angle disrupts reliable indirect sequential navigation,
while the biased diffusive agents remain robust (\ref{fig:noise_visangle} Fig).
Taken together, the agents traverse perception-action attractors approximately centered on the patch,
structured via the same invariances perceived by the indirect sequential agents, 
yet their movement pattern differentiates how they are used.

\subsection*{Angle-only navigation: classification metrics}

% robust sep - relative ori x spatial/histo
While the behavioral bifurcation between the two angle-based navigation strategies is qualitatively apparent,
robust separability calls for quantified measures.
General movement behavior can be compared by relating orientation to initial heading as an agent approaches the patch
(Fig~\ref{fig:trajs_corrs}a/b, bottom row),
demonstrating the tendency for straight, sparse or spinning, disperse profiles.
However, though these plots provide useful intuition, 
extracting robust separation between the two classes for the entire range of agents has proven to be difficult,
as can be seen in the extended figure 
(\ref{fig:corrs_ext}a/b Fig, bottom row, top plots).

% temporal - decorrelation - chemotaxis
Instead of tracking relative orientation in space, its temporal counterpart provides a cleaner split 
(Fig~\ref{fig:trajs_corrs} bottom row, bottom left).
Although other differences can be observed, 
the time needed for half of the agents to decorrelate from their initial heading
consistently differentiates the two agent types.
Biased diffusive agents quickly turn away, 
whereas indirect sequential agents follow their initial heading longer, until reaching an elliptical manifold.

% spatial - directedness
Another way to track spatial correlation is via directedness, 
described as the orientational predictability of an agent at a given point in space.
Average directedness is evidently lower for the biased diffusive agents
(Fig~\ref{fig:trajs_corrs} bottom row, bottom right).
The measure quantifies observations made for the top correlation plots of Fig~\ref{fig:trajs_corrs}:
indirect sequential agents tend to stick to few directions, while biased diffusive agents are more directionally indiscriminate.
Yet for the former, directedness is locally low near elliptical manifolds and the patch, 
reflecting the wider range of potential directions at defined decision points,
whereas there are no consistent spatial structures seen for the latter.

\begin{figure}[htb]
	\centering
	\includegraphics[width=\linewidth,trim={0 0 0 0},clip]{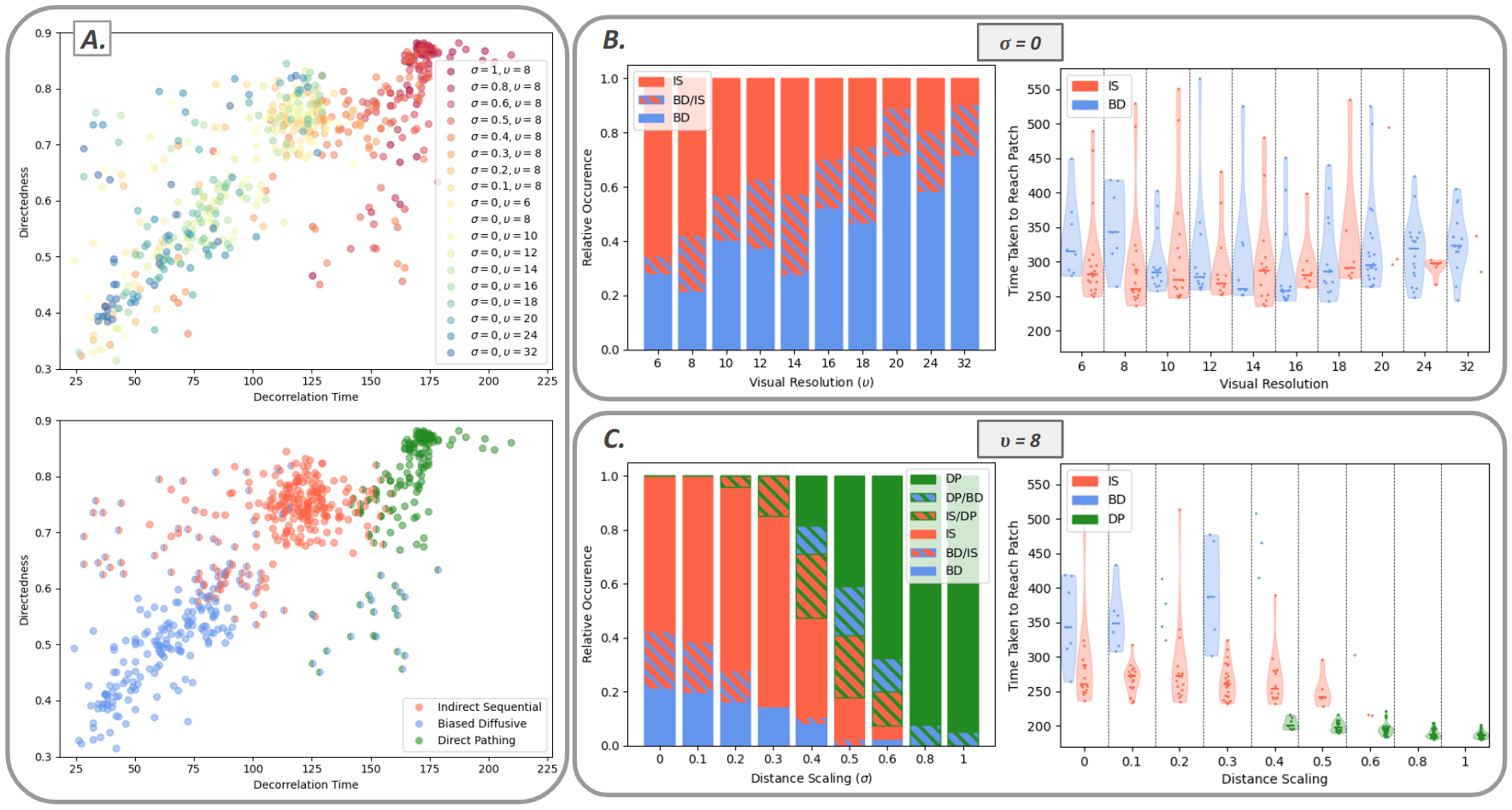}
	\caption{
	  \textbf{Classification \& relative evolvability and fitnesses.}
	  A: Navigational algorithmic classes, separated by decorrelation time and directedness,
	  top: colors correspond to distance scaling factor ($\sigma$) and visual resolution ($\upsilon$),
	  bottom: colors correspond to algorithmic classes (IS: indirect sequential, BD: biased diffusion, DP: direct pathing), including hybrids as combined half-circles.
	  B: Angle-based strategy bifurcation,
	  left: relative rates of either angle-based classes evolving under different visual resolutions,
	  right: relative fitnesses, revealing optima at $\upsilon = 8$ for IS and $\upsilon = 16$ for BD.
	  C: Distance-based phase transition, similar plots as in B, varying distance scaling factors and $\upsilon = 8$.
	  Scatter plots on right include violins for sample size greater than 5.
	}
	\label{fig:classes}
  \end{figure}

\subsection*{Angle-only navigation: separation \& evolvability}

% decorr x direct - separation - fitness attractor
Two clusters emerge in the phase space of decorrelation time and directedness, reflecting the qualitative separation observed above
(Fig~\ref{fig:classes}a, \ref{fig:classes_extended} top left).
How such clustering arises can be seen by plotting average fitness values for each agent or run
(\ref{fig:classes_extended} Fig, top right).
Runs close to cluster centers tend toward the greatest fitness, while those farther tend to drop.
Such gradients suggest the two clusters are attractors in a fitness landscape, where optimization pulls toward the centers.

% t-SNE/UMAP
Nonlinear dimensionality reduction techniques can find the gradient between the two navigational classes
(\ref{fig:classes_dimred} Fig)
directly via the set of all possible visual observations and actions in the environment,
however with lower explainability than using the two correlation metrics.

% vis_res modulation --> evolvability + fitness
Given the asymmetric effect of noise to visual angle, it was expected that visual resolution 
would be critical in regulating preferential evolvability between the two classes.
Indeed it is (Fig~\ref{fig:classes}b): 
lower visual resolution favors learning an indirect sequential strategy, while higher favors biased diffusive.
Whereas the indirect sequential mechanism involves distinguishing only few major transitions,
biased diffusive perception needs to construct an accurate gradient with respect 
to the entire range of potential visual observations, locations, and orientations.
Therefore, visual resolution, a parameter dictating the degree of environmental complexity able to be perceived,
affects relative occurence between strategies accordingly.

% vis_res x fitness
Relative fitness with respect to visual resolution shows a similar pattern
(Fig~\ref{fig:classes}b).
Each navigational strategy confers similar optimal global performance, explaining why the behavioral bifurcation exists in this parameter space.
However, visual resolution needed for the biased diffusive agent to meet this level is twice that of the indirect sequential,
thus requiring twice the number of convolutional operations per timestep, 
as well as twice the complexity in the visuospatial affordance space
(\ref{fig:map_rays} Fig, 4th row).

\subsection*{Navigation with distance}

Adding distance information to the visual input gives rise to a third class of algorithms that can travel directly to the patch
(Fig~\ref{fig:trajs_corrs}c).
Each agent differs in the degree and arrangement of how common routes collapse as they approach the patch, 
though variation is slim compared with the two angle-based classes.
The uniform directness of their routes is apparent in their relative heading profiles,
long decorrelation times, and high directedness throughout the environment, with the exception of the patch area.
Appropriately, their neural representations appear to outline clear directional trajectories to the patch position
(Fig~\ref{fig:neureps}C/F)
with oriented activations cleanly situated between environment boundaries and the patch.
% However, although the patch position is clearly outlined, its activation does not illustrate the spatial sparsity of a place cell
% (Fig~\ref{fig:neureps} left).
The goal position is outlined using either dataset, with strong orientational alignment when using grid data, 
appearing to represent a goal-direction cell~\cite{sarelVectorialRepresentationSpatial2017}
as well as behaviorally affording goal vectors regardless of initialization
(Fig~\ref{fig:neureps}F, middle).

Distance-based navigation establishes a third cluster center in the correlation space, 
(Fig~\ref{fig:classes}a).
due to the lack of behavioral variability possible, whereas many different movement patterns can be classified under either angle-based class.
As the distance signal is attenuated, relative occurrence of direct pathing agents decreases until the other two classes fully occupy the resulting space
(Fig~\ref{fig:classes}c left).
Given that a sizable fitness difference between distance-based and angle-based strategies remains even with attenuated signal
(Fig~\ref{fig:classes}c right),
the abrupt disappearance of direct pathing behavior suggests the networks are no longer able to effectively use the input.

\section{Discussion}

% \subsection{Emergent Navigational Algorithms}

% overview
We trained a simple yet expressive artificial neural network 
to navigate toward a hidden patch in a robust and efficient manner 
(Fig~\ref{fig:flow}, \ref{fig:evol_params}).
By placing heavy constraints on perception, cognition, movement, and the environment, 
that is, 
monocular 1D visual perception,
no predictive framework or within-trial memory,
a single decision variable governing movement kinematics (turning \& speed),
% and a highly degenerate environment in percept-state space,
and a four wall or landmark environment,
we demonstrate the sufficiency of an egocentric, feedforward, route-based framework
to solve the Morris Water Maze.

% emergent algos - behavior + mechanism
Three distinct classes of algorithms emerged in order to accomplish this, each with a set of distinguishable features
(Fig~\ref{fig:trajs_corrs}, \ref{fig:neureps}).
The first class of agents navigate by composing a sequence of straight segments, indirectly oriented with respect to the patch, 
with major turns defined by elliptical-shaped manifolds, i.e. view-specific invariances stemming from perception of two adjacent corners.
The second class instead learns to spin or diffuse toward the patch,
driven by smoother global gradients constructed from the same elliptical invariances.
And the third uses distance perception, a phenomenological implementation of advanced perceptual ability, 
to calculate near-ideal paths to the patch.
Described via behavioral characteristics, these classes are termed indirect sequential, biased diffusive, and direct pathing, respectively.

% emergent algos - classification
The classes can be separated by two correlation metrics: decorrelation time and directedness
(Fig~\ref{fig:trajs_corrs}), 
indicating temporal persistence in heading and directional potential in space.
Biased diffusive has low measures for both, indirect sequential has medium, and direct pathing has high.
The three classes tend to cluster into distinct bins in this 2D phase space, 
suggesting fitness-based attractors for movement correlation parameters.

% emergent algos - evolvability + vis_res / dist_scal
Angle-based algorithms (indirect sequential and biased diffusive) 
confer comparable performance though prefer different visual resolutions
(Fig~\ref{fig:classes}).
Indirect sequential agents evolve more readily and perform better with lower visual resolution and biased diffusive with higher.
While only requiring low quality visual signal and sparse orientational profiles, 
the indirect sequential agents are highly susceptible to visual angle noise throwing them off-course,
whereas such noise does not affect the dynamic, disperse process of biased diffusive agents.
Distance-based, direct pathing agents perform better than either angle-based class, 
resulting in an abrupt phase transition as the distance input becomes computationally usable,
although requiring another dimension to their perception.

\subsection*{Vision vs. vectors}

% map-route asymmetry
Our understanding of route-based strategies lags behind that of cognitive maps,
in part due to historical preference for the latter
\cite{hartDevelopmentSpatialCognition1973, siegelDevelopmentSpatialRepresentations1975, 
warrenNonEuclideanNavigation2019, wangEgocentricAllocentricRepresentations2020, kunzNeuralCodeEgocentric2021}.
Predispositions toward maps persist via experimental paradigms that reinforce the primacy of Euclidean graphs, 
despite partial dissociability and parallel use of the two strategies,
and even despite results demonstrating similar rates of navigational success with either strategy
\cite{marchetteCognitiveMappersCreatures2011, sheltonChapterSixMechanistic2013, weisbergCognitiveMapsPeople2018, goodroeComplexNatureHippocampalStriatal2018}.
Whereas calculating shortcuts or direct trajectories are certainly optimal in certain scenarios,
overemphasis on cognitive maps denies both value and understanding of route-based systems.

% comp complexity + env context
Given that map-based knowledge demands construction of a recurrent internal world model to predict and integrate external dynamics,
complexity required presumably far surpasses systems that directly respond to visual input.
Assuming a given organism would seek to minimize computational complexity while maintaining successful performance,
external factors such as environmental context play a critical role in determining which system preferentially develops.
For example, path integration may be ideal for desert ants which require precision to quickly return home after foraging in midday heat 
\cite{wehnerParallelEvolutionThermophilia2011},
though may cause problems in unpredictable areas with potentially dangerous shortcuts.
Route-based representations, in contrast, have been suggested to favor dense environments 
with many navigational constraints or without ambiguous landmarks such as jungles or mazes
\cite{babayanHippocampocerebellarCentredNetwork2017, peerStructuringKnowledgeCognitive2021, spiersHowEnvironmentShapes2022}.
What has not been evidently studied is their sufficiency in open arenas, an environment suggested to favor cognitive maps
\cite{kesslerHumanNavigationStrategies2024}.
Our results present non-intuitive evidence:
navigational strategies based on visual routes can perform well in open arenas.
% A bottom-up approach aiming at identifying minimally sufficient processes for effective behavior remains valuable for providing 
% the complementary perspective to studying range in potential capability.

% RL: model-based vs model-free / goal-directed vs habits --> not quite map vs response
The reinforcement learning community represents the navigational dipole somewhat differently.
Model-based learning entails planning upon a predictive, goal-directed framework by constructing a model of external dynamics,
while model-free works directly on cached action-value pairs
\cite{dawUncertaintybasedCompetitionPrefrontal2005, woodPsychologyHabit2016}.
Described in this way, model-based learning appears to reflect map-based strategies 
and model-free to response sequences, coordinated through associative memory,
\cite{babayanHippocampocerebellarCentredNetwork2017, anggrainiNeuralSignaturesReinforcement2018},
though the two perspectives may not fully align
\cite{khamassiIntegratingCorticolimbicbasalGanglia2012, herwegSpatialRepresentationsHuman2018, vijayabaskaranNavigationTaskAction2022}.
Despite discrepancies, balanced theoretical and experimental effort toward both model-based and model-free reveals contextual advantages for each.
Model-based learning tends to be more efficient in learning, as well as flexible and theoretically effective in control, 
yet require significantly higher capacity to handle complex external dynamics, which in turn may lead to overfitting
\cite{nagabandiNeuralNetworkDynamics2018, kimTaskComplexityInteracts2019}. % feinberg
In contrast, model-free learning is computationally cheap and effective at scale.

\subsection*{Neural connections}

% distributed neuro systems
The hippocampus and striatum are typically contrasted with respect to the Morris Water Maze, 
where the former is responsible for map-based, allocentric strategy with a focus on environment boundaries and distal (extra-maze) landmarks,
and the latter for response-based, egocentric strategy using proximal (intra-maze) landmarks
\cite{doellerParallelStriatalHippocampal2008, goodroeComplexNatureHippocampalStriatal2018, parra-barreroMapSpatialNavigation2023}.
One possibility not examined in this dual-solution focus is the construction of egocentric routes based on boundaries and distal landmarks
\cite{harveyEmergenceEgocentricCue2008}.
Accounting for this third option, i.e. equating routes as sequences of decisions, may help unite seemingly disparate functions of the hippocampus:
spatial navigation and episodic memory
\cite{goodroeComplexNatureHippocampalStriatal2018, wangEgocentricCodingExternal2018, wangEgocentricAllocentricRepresentations2020, 
kunzNeuralCodeEgocentric2021, mooreLinkingHippocampalMultiplexed2021, parra-barreroMapSpatialNavigation2023}.
Rodents with hippocampal lesions are confined to beaconing, thigmotaxis, and scanning behavior
\cite{riceLesionsHippocampusDorsolateral2015, higakiRecognitionEarlyStage2018, curdtSearchStrategyAnalysis2022, parra-barreroMapSpatialNavigation2023},
while this can be understood as inflexible striatal navigation, it does not relate to the robust, initialization-invariant routes seen in our trained agents.
The neural-semantic dipole between hippocampal-maps and striatal-responses does not leave room to allow for flexible route-based navigation,
which presumably uses hippocampal episodic sequences to navigate
via both egocentric responses and allocentric boundary or distal information,
yet without demanding the ability to localize or plan on a map-like substrate.
% Contrasting this hypothesis, rodents born without a hippocampus have been shown to flexibly compose navigation sequences in a constrained maze task
% \cite{zhengMiceManhattanMaze2024},
% though genetic manipulation may simply allow transfer of hippocampal-like calculations onto other structures during development.
% For this paper, we use 'route-based' to describe this third option, 
% since although 'guidance' or 'triangulation' may also describe our agent behavior
% \cite{parra-barreroMapSpatialNavigation2023},
% these terms do not have widespread use or even mutual agreement in the navigation community.
%% Note - episodic (memory) could be downstream/complementary to episodic (vis-action sequences)

% sensory + cognitive + experimental degeneracies
The distributed nature of these systems generates degeneracies on multiple fronts in the solution space,
allowing an organism to choose which pathway might suit the context.
Outside the hippocampal-striatal dipole, several other cortical areas process navigational information, including: 
the retrosplenial cortex~\cite{augerRetrosplenialCortexCodes2012}, 
occiptal parietal cortex~\cite{bonnerCodingNavigationalAffordances2017}, 
and even visual cortex~\cite{saleemCoherentEncodingSubjective2018}.
As described in the introduction, two recent models that allow for visual information,
as well as vestibular~\cite{baninoVectorbasedNavigationUsing2018} or distance~\cite{recanatesiPredictiveLearningNetwork2021},
did not learn visual dependencies.
While the former two information streams and learned representations are sufficient for the task, 
they are incomplete on their own
\cite{parra-barreroMapSpatialNavigation2023, krausseHeaddirectionSystemShows2025}.
Reflected in neural architecture, two distinct sets of head direction cells are suggested to be updated via either self-motion or visual cues
\cite{jacobIndependentLandmarkdominatedHeaddirection2017, kornienkoNonrhythmicHeaddirectionCells2018},
though see also evidence of integration \cite{siegenthalerVisualObjectsRefine2025}.
Guidance and aiming tasks have been shown to prefer allocentric and egocentric action spaces, respectively, 
though both can be solved with the other regardless~\cite{vijayabaskaranNavigationTaskAction2022}.
An important point not often discussed, let alone remedied, 
the Morris Water Maze does not adequately separate guidance from map navigation~\cite{parra-barreroMapSpatialNavigation2023}:
since landmarks are available from every point in the environment, 
triangulation via visual angles may be easier than self-localizing and computing vectors on a coordinate system 
\cite{wolbersChallengesIdentifyingNeural2014, eichenbaumRoleHippocampusNavigation2017}.
Degeneracy in the cognitive, sensory, action, and task spaces affords no easy answers.
One way to inspect such systems is to focus on silencing or perturbing parts of these pathways
\cite{riceLesionsHippocampusDorsolateral2015, aghajanImpairedSpatialSelectivity2015, acharyaCausalInfluenceVisual2016, 
higakiRecognitionEarlyStage2018, mooreLinkingHippocampalMultiplexed2021, vijayabaskaranNavigationTaskAction2022},
and another, used here, is to explore a simple model constrained along one particular, under-studied pathway.

% specific neural correlates
But does the agent learn a map?
The agent learns to approximate a function that transforms egocentric observations into useful actions.
Learned neural representations show significant directional modulation (Fig~\ref{fig:neureps}),
aligning with a flight of papers on rodents in VR
\cite{aghajanImpairedSpatialSelectivity2015, acharyaCausalInfluenceVisual2016, mooreLinkingHippocampalMultiplexed2021}
and a modeling study examining task and action constraints
\cite{vijayabaskaranNavigationTaskAction2022}.
Neural tuning corresponds with patch location,
mirroring reports of representations with respect to goals
\cite{sarelVectorialRepresentationSpatial2017, ormondHippocampalPlaceCells2022},
as well as boundaries and landmarks
\cite{wangEgocentricCodingExternal2018, alexanderEgocentricBoundaryVector2020, siegenthalerVisualObjectsRefine2025}.
The elliptical manifolds of the indirect sequential strategy trace view-specific correlations widely observed in primates and humans
\cite{rollsSpatialViewCells1997, ekstromCellularNetworksUnderlying2003, jacobsSenseDirectionHuman2010, maoSpatialModulationHippocampal2021, pizaPrimacyVisionShapes2024},
which may be linked to egocentric cells that conjunctively encode distance
\cite{kunzNeuralCodeEgocentric2021}.
While we do not rule out the possibility of adapting these learned visuospatial functions to a map where the agent can plan trajectories,
through the architectural constraints of our model, we have demonstrated that this externalization is not necessary.
Similar to a minimal model by Bastien \& Romanczuk proving vision alone can explain collective behavior without the need for higher level representations
\cite{bastienModelCollectiveBehavior2020},
and echoed by studies on sensorimotor contingencies and enactivist robotics
\cite{hoffmannRobotsPowerfulAllies2018, degenaarSensorimotorTheoryEnactivism2017, egbertUsingEnactiveRobotics2022},
the main point of our model is to show that directly perceived routes provide sufficient navigation without requiring map-based inference.

% behavioral analysis complements neural
Moreover, correlating average neural activity with the external environment may mislead.
The averaging of neural activity of our evolved agents over space and time reveals activity patterns correlated with goal direction
(Fig~\ref{fig:neureps}A/B, right).
However, a corresponding goal signal is absent in time-resolved activation patterns governing actual movement decisions of our agents
(Fig~\ref{fig:neureps}D/E, right).
Thus, the general question emerges whether empirically observed, averaged activation patterns correlating with higher-order task features 
are indeed involved in decision processes in real-time, or whether they may emerge as a "by-product" of learning the task via low-level decision rules
(e.g. indirect sequential) and a corresponding averaging procedure.
Or to put it in other words: to what extent are map-based representations, extracted through averaged activity patterns, directly involved in the generation of behavior, 
rather than a necessary statistical consequence of a complex neuronal system learning a spatial task?
A recent studies shows how neural activity describes internal action sequences rather than external variables
\cite{zutshiHippocampalNeuronalActivity2024},
which may help explain why it is difficult for us to integrate between routes
\cite{weisbergVariationsCognitiveMaps2014, warrenNonEuclideanNavigation2019}.
Characterizing behavior provides a complementary perspective to mean field neural analysis
\cite{harveyEmergenceEgocentricCue2008}, which may even be critical to understand direct perception-action behavior.
While directional modulation may describe average neural correlation, 
clockwise curving trajectories may indicate a biased diffusive navigational strategy
\cite{mooreLinkingHippocampalMultiplexed2021}
or attractor-like alignment may indicate the use of visual manifolds
\cite{vijayabaskaranNavigationTaskAction2022}.

\subsection*{Behavioral convergence}

% MWM
Rodent paths observed in the Morris Water Maze reflect movement characteristics from each of the three classes in the present study.
While many learn how to take optimal direct paths, others travel in clearly wrong directions before abruptly turning toward the patch
\cite{cookePathfinderOpenSource2020, curdtSearchStrategyAnalysis2022, villarreal-silvaAgedRatsLearn2022}. 
Although often termed "indirect search", this pattern parallels the indirect sequential navigational strategy.
Other rodents instead walk toward the patch yet spin as they approach, 
attributed as "self-orienting" or even "nonsense movements" of "directed search"
\cite{grazianoAutomaticRecognitionExplorative2003, cookePathfinderOpenSource2020, curdtSearchStrategyAnalysis2022, villarreal-silvaAgedRatsLearn2022},
they mirror biased diffusive trajectories.
Shifting perspective to view such behavior as essential components to egocentric navigational routes, 
as opposed to search behavior or errors relative to ideal trajectories, may yield novel hypotheses.
Perhaps these rodents prioritize angular over distance information
\cite{etienneNavigationSmallMammal1992, wallaceRatsMaintainOverhead2013, hagbiHeightsPlainsHow2022}.
Balanced relative occurence metrics and across-trial stability of the three strategies
\cite{grazianoAutomaticRecognitionExplorative2003, cookePathfinderOpenSource2020}
suggest each may confer contextual tradeoffs, where diversity has evolved to ensure robust population-level fitness in a variable environment.
% In other words, there may be evolutionary benefit to angle-based behavior, which may be better understood as a navigational strategy than as error.
% While formalizing these comparisons are possible, in the interest of minimizing narrative complexity, it is left for future work.

% sperm chemotax
The biased diffusive trajectories appear to mirror the helices of sperm cell chemotaxis
\cite{alvarezComputationalSpermCell2014, rodeInformationTheoryChemotactic2024}. % ramirez-gomez, zaferani
Sperm cells modulate swimming path curvature via temporally sampled chemical concentrations to bias motion up-gradient towards an egg.
Despite clear differences in perception and movement capabilities, 
both learn a similar movement strategy to robustly navigate to a hidden target.
Though clearly the mechanisms differ, as sperms cannot see, the higher level phenomena seem to overlap,
suggesting fundamental algorithmic properties for gradient navigation, which may be either directly sampled or indirectly constructed via angle-based vision.

% similar elliptical geometry
Elliptical decision-making thresholds have been recently found to govern spatial bifurcations between choices or goal vectors
\cite{sridharGeometryDecisionmakingIndividuals2021, gorbonosGeometricalStructureBifurcations2024}, 
in the same way ellipses spanning adjacent corners govern turn decisions for the indirect sequential agents.
Fruit flies, desert locusts, and zebrafish have each been observed to move toward the average of two choices until reaching a critical threshold, 
spontaneously forcing a decision towards one or the other choice, even in sequence if more than two are available.
While in the current study, perception-action loops learned to deterministically use elliptical thresholds to compose route segments,
here they were used by a ring attractor model to stochastically choose between goals or sets of goals.
Despite differing computational, algorithmic, and implementational level details, 
resulting behavior appears to have converged on a fundamental element:
angle-based or view-specific turn decisions with respect to two spatially distinct entities,
proposing a non-Euclidean spatial representation.

% discrepancies - architecture, optimization, 
While there are speculative comparisons to make between our model and biology, there are limitations.
The model is intended to be phenomenological, thus only general relations should be made.
Although our convolutional neural network may reflect the visual cortex
\cite{lindsayConvolutionalNeuralNetworks2021},
its operations as well as the downstream layers are meant to
ambiguously define sensorimotor perception-action information flow rather than specific brain regions or operations.
Complexity has been reduced to a simple feedforward network, whereas the visual cortex and associated structures contain recurrent connections.
Similarly, neural network optimization accumulates thousands of parallel perturbations and experiences each round, 
and thus is intended to reflect neither the speed nor mechanism of biological learning,
nor does it reflect the search process preceding navigation.
Only the behavior of a fully trained network is intended to be related to behavior of fully trained animals or humans.

\subsection*{Shifting perspectives}

% constraint approach
Rather than assuming the entire range of perceptual and cognitive capabilities of organisms when modeling behavior,
probing minimal sufficient conditions for successful behavior under energetic-bound constraints
offers unique insight into how organisms may behave outside the lab.
Though many organisms may have the ability to construct predictive frameworks or integrate vestibular cues,
a simpler feedforward system might do just fine, without the energy expenditure.
That the resulting behavior and neural correlates track with phenomena covering rodents, insects, fish, and primates,
provides evidence that such constraints may be widely applicable.
These simple models do not assume navigation behavior to solely rely on one of the observed strategies,
but rather suggest low dimensional visual techniques that can be used to save memory or complexity.

% amicability clause
This is not intended to argue against the importance for cognitive maps, distance perception and prediction, or vestibular integration,
nor to deny crossmodal interactions between map and route pathways,
but rather encourage a plurality of perspectives moving forward. 
Given our historical bias toward top-down representation, 
as well as our physical experience using maps or viewing rodents in a maze beneath us,
we need to approach the egocentric, route-based perspective with a new lens.
Route-based navigation can provide a fully separable navigational strategy with its own mechanisms and logic, 
demanding novel experimental methods for disentangling systems and characterizing mechanisms,
proposing alternative navigational hypotheses to contend with assumptions of search or errors. 
Each system has characteristic advantages and disadvantages, 
with evolution and development shifting the balance toward a particular direction in a given environmental context.
Only by understanding both can we more fully grasp the breadth of tools potentially available to a navigating organism.

\section{Materials and methods}

\subsection*{Visual encoding}

Vision is simulated as a raycast originating from a single retina at the front edge of the agent,
the orientation of both visual retina and agent are bound as a single state variable.
Perceptual input is structured as a 1D array with each element one-hot encoded with respect to wall identity.
Though a simplification from 2D vision, collapsing the vertical dimension has been
previously found to retain sufficient directional information for visual navigation
\cite{wystrachHowFieldView2016, zeilVisualNavigationProperties2023}.

Two key hyperparameters govern the raycasting process: field of vision (FOV) and resolution.
FOV stretches the limits of the raycast with respect to the agent, and resolution determines the number of rays to cast within these limits.
The FOV is set at 40\% or 144 degrees to mirror the functional visual field for humans, as measured by reaction time to discrimination tasks
\cite{sandersAspectsSelectiveProcess1970}, though the impact of FOV on navigation performance was also assessed 
(\ref{fig:evol_params} Fig, 1st row).
The visual resolution ($\upsilon$) is minimally initialized at 8 rays for much of the analysis and later varied to examine its effect on behavior
(\ref{fig:evol_params} Fig, 2nd row).

Weber-Fechner logarithmic scaling is applied for distance information, similar to how height is perceived by our eyes~\cite{buzsakiLogdynamicBrainHow2014}.
The scaling is calculated as follows:

\begin{equation}
  y = - {\frac {1}{k(\sigma)}} \ln(x) + m(\sigma)
\end{equation}

\noindent\ with $x$ representing distance from the perceived environmental boundary, 
and $k$ and $m$ fixed according to the distance scaling factor $\sigma$,
by relating minimum and maximum possible visual distances to a fractional visual encoding range.
This difference in distance results in maximal encoding variance at $\sigma = 1$, 
and to 10\% of the encoding space $\sigma = 0.1$.

\subsection*{Environment}

The arena is defined at 1000 x 1000 units and the patch at 50 units.
The patch to arena ratio correlates a 1-2 m diameter arena and 10-15 cm diameter platform often used for rodents in the Morris Water Maze
\cite{grazianoAutomaticRecognitionExplorative2003, cookePathfinderOpenSource2020, curdtSearchStrategyAnalysis2022, villarreal-silvaAgedRatsLearn2022, baninoVectorbasedNavigationUsing2018}.
An arbitrary timescale was chosen for our simulations, with a maximum agent speed at 2 units per timestep.
% This may correspond either to 15 cm/s swimming speed, a 3.75 m patch radius, and a 75 x 75 m arena for rodents,
% or 1.5 m/s walking speed, a 37.5 m patch radius, and a 750 x 750 m arena for humans.
This could be shifted to describe a lower or greater decision and action speed to reflect rodents, humans, or other organisms.

% % Assuming an average laboratory environment with a width of about 5-15 m, 
% % the simulated boundary corners may be classified as exceptionally distal landmarks, 
% % and the environment as a very open arena.
% Arena boundary corners are classified as distal landmarks
% Our focus on distal landmarks runs contrary to the algorithm trained by Banino et al,
% where ablation studies demonstrate such landmarks to be functionally irrelevant
% \cite{baninoVectorbasedNavigationUsing2018}.

Environmental geometry provides the spatial reference needed for both orientation and location~\cite{chanObjectsLandmarksFunction2012}.
Although landmark and geometric encoding have been argued to be performed by separate modules 
\cite{wangHumanSpatialRepresentation2002, leeTwoSystemsSpatial2010, doellerDistinctErrorcorrectingIncidental2008, doellerParallelStriatalHippocampal2008}, 
in the present study the two are not differentiated.

\subsection*{Network architecture}

The CNN used is a simplified version of the ConvNeXt v2 architecture~\cite{liuConvNet2020s2022, wooConvNeXtV2CoDesigning2023},
a state-of-the-art design that outperforms earlier varitions in the CNN design space and rivals the best vision transformers.
While simpler designs exist, the ConvNeXt was chosen for its separable depthwise and pointwise convolutions, 
where separate parameters act on the orthogonal depth and channel axes, enabling greater expressibility for relatively low additional computational cost.
A transformer was not chosen since they lack biological inductive biases, such as translation equivariance,
and due to their novelty, lack the understanding and tools currently available for CNNs~\cite{khanTransformersVisionSurvey2022}.

Beyond critical minimal values, depth and dimension of the CNN and perceptron did not noticeably affect performance 
(\ref{fig:evol_params} Fig, 3rd/4th rows).
Notably, two channel outputs for the CNN is insufficient for agents to reliably solve the task,
possibly reflecting an intrinsic, minimal dimensionality to the navigation problem
\cite{liMeasuringIntrinsicDimension2018}.
For more complex and recurrent architectures, as well as those with proprioceptive feedback,
data at adequate sample size is forthcoming at the time of submittal and will be appended upon review.
The activation function used for the CNN is a rectified linear unit.
A sigmoid linear unit was found to net marginal performance increases, though slower simulation times.

The number of parameters is less than 300 for the models presented in the main text,
significantly less than the tens of thousands needed for the predictive navigation network used in~\cite{recanatesiPredictiveLearningNetwork2021},
and still less than the thousand needed for the non-visual linearized predictive networks in~\cite{stocklLocalPredictionlearningHighdimensional2024}.

\subsection*{Agent movement}

Network output is a single continuous value that directly represents turning angle for the next timestep,
bounded by \ang{90} left to \ang{90} right via a hyperbolic tangent function.
Speed is scaled by the output via a linear function, with zero representing maximum speed and both left and right limits for when the agent is stopped.

Collision with the foraging patch results in simulation termination when training, but no effect when testing.
Collision with the boundary stops the agent from traveling out of the environment but allows rotation, 
it does not result in repulsive physical effects or a change in sensory input.

\subsection*{Optimization}

The ES algorithm used is Policy Gradient Parameter Exploration (PGPE) using the ClipUp optimizer 
\cite{sehnkeParameterexploringPolicyGradients2010, tokluClipUpSimplePowerful2020},
with parameters outlined in Table \ref{EA_params}.
Among various ES algorithms, PGPE was chosen for its balance between performance and speed, 
with greater performance than OpenAI-ES and greater speed than CMA-ES~\cite{haVisualGuideEvolution2017}.
ES entails less complexity than RL, as there is no need for differentiability, value approximation, or within-trial credit assignment 
\cite{salimansEvolutionStrategiesScalable2017, majidDeepReinforcementLearning2021}.
The main drawback of ES, sample inefficiency, is acceptable given its parallelizability.

\begin{table}[h]
	\centering
	\renewcommand\arraystretch{1.2}
	\caption{\textbf{Hyperparameters used for the ES algorithm}}\label{EA_params}
	\begin{tabular}{@{}cc@{}}
	\toprule
	Name     & Value   \\
	\midrule
	Generations       & 1000   \\
	Episodes          & 20   \\
	Population Size   & 50   \\
	Standard Deviation, Initial   & 0.1   \\
	Standard Deviation, Learning Rate   & 0.1   \\
	Standard Deviation, Max Change   & 0.2   \\
	Mean, Learning Rate   & 0.2   \\
	ClipUp, Momentum   & 0.8   \\
	ClipUp, Max Speed   & 0.4   \\
	\bottomrule
	\end{tabular}
  \end{table}

\subsection*{Training, validating, testing}

Fitness performance during training was calculated as the time taken to reach the patch plus remaining distance if the agent has not reached the patch within simulation time limit (1000 timesteps).
Alternative scalings were tried but found to not make a significant difference.

Validation was performed to more robustly test performance from a wider range of initializations than the episodes during a training generation.
Twenty of the top performing generations were chosen, using the population parameter center for each, with 100 initializations for each.
Remaining distance from the patch was not included for the validation data.

The agent was tested by simulating the top agent of an evolutionary run at different intializations across the grid, 
spaced every 25 units and oriented in 16 directions.
Each initialization was run for 500 timesteps, regardless if the agent collides with the patch.
The test data from this set of trajectories was used for Fig~\ref{fig:trajs_corrs}.

\subsection*{Elliptical manifold perception}

How the elliptical manifolds, 
explicit for the indirect sequential agents (Fig~\ref{fig:trajs_corrs}a)
and implicit for the biased diffusive agents (\ref{fig:traj_vecfield_ori} Fig),
can be examined directly via perturbations. 
Perceptual, environmental, and movement related parameters all affect the shape and location of the manifolds
(\ref{fig:perturbs} Fig).

Inferring the generative mechanism,
given the ellipses emanate from two adjacent corners, 
the manifolds may be marked by two raycasts simultaneously intersecting both corner locations,
representing a change in wall perception in both corners at once.
A numerical search for these ideal locations shows that simultaneous dual corner detection reconstructs elliptical arcs,
with orientation rotated about the center between the two corners,
and both perturbed FOV and visual resolution shifting the arcs predictably
(\ref{fig:map_rays} Fig, 1st \& 4th rows).

While perceptual and environmental parameters are fixed for a given training period, movement dynamics can shift manifold positioning as needed.
Generated ideal elliptical arcs intersect with patch position at FOV of 40\%, but not for 35 or 45.
Though when trained with these FOVs, the agents learn to approximate ideal arcs useful for calculating patch position regardless
(\ref{fig:map_rays} Fig, 2nd row).
By increasing turning speed before intersecting with a manifold, the agent expands its capacity to perceive a corner, 
thereby allowing the agent to effectively precede ideal arc location.
Looking at turning speed for each agent, it is clear that the agents with FOV 35 or 35\% 
increase incoming turning speed in order to shift the ellipse to a more useful position
(\ref{fig:map_rays} Fig, 3rd row).
Thus, although various perceptual, action, and environmental parameters constrain elliptical manifold construction,
the learning algorithm coordinates movement behavior such that generated ellipses facilitate navigational success.

Furthermore, mean field analysis can demonstrate how the two angle-based classes use the elliptical manifolds differently
(\ref{fig:traj_vecfield_ori} Fig).
Indirect sequential agents (a) can sufficiently navigate without minima overlapping patch location
since their algorithm relies on a sparse, stable subsample of orientations.
Biased diffusive agents (b) navigate with respect to a disperse, dynamic sample of every orientation,
thus their success depends on correlating mean field minima with patch location.
Elliptical manifolds visibly structure attractors for both angle-based classes,
and are not apparent with the smoother gradients of distance-based agents (c).
Higher visual resolution angle-based individuals (bottom row, a/b) 
tend to exhibit increased elliptical manifold presence (south \& east walls),
possibly explaining why biased diffusive agents perform better at greater visual resolution.

\subsection*{Movement correlations}

All movement correlations were transformations of the test trajectory data previously described.
The first 25 timesteps, when the agent is orienting itself upon random initialization,
are cut out when calculating movement statistics. 
Thus, the orientaion at t=25 represents the initial route heading.

The spatial heading profiles of the top left correlation plots in Fig~\ref{fig:trajs_corrs} 
were produced by plotting agent orientation relative to the initial route heading against the distance from the patch center.
The polar histograms count the frequency of relative orientations, where the area under each bar is scaled to relative frequency.

The temporal correlation metric uses a function borrowed from a sperm chemotaxis paper~\cite{armonTestingHumanSperm2012}, 
given the similarity between movement trajectory patterns:

\begin{equation}
  C(t) = <\cos[\phi(t _{0} + t) - \phi(t _{0})]>
\end{equation}

\noindent\ with $\phi$ representing the orientation angle of the agent and $t _{0}$ defined at a delay of 25 timesteps from initiation. 
Decorrelation time is defined by the length of time needed for $C{t}$ to drop below half.

Directedness, the spatial correlation metric, is based on the range of directional possibilities available to the agent, mapped across space.
We calculate the information entropy 
\cite{shannonMathematicalTheoryCommunication1949, pilkiewiczDecodingCollectiveCommunications2020} 
of the orientational frequency distributions for each spatial bin across the environment:

\begin{equation}
  H(\Phi,b) = -\sum_{\phi \in \Phi}p(\phi)\log(p(\phi))
\end{equation}

\noindent\ with $H$ being the directional entropy, $\Phi$ being the set of 16 orientations, and $b$ the spatial bin.
Bounding to the range of entropy values possible for $\Phi$ and inverting, we arrive at a measure of directedness:

\begin{equation}
  D(H,\Phi,b) = \frac{H_{max}(\Phi) - H(\Phi,b)}{H_{max}(\Phi) - H_{min}(\Phi)}
\end{equation}

\noindent\ Given the initial delay and that trajectories generally move toward the center of the environment, the edges of the environment contain lower data density,
thus data 100 spatial units away from the boundaries were masked out of the calculation of directedness.
Trajectory information near the patch is external to navigation behavior, 
thus data 100 units from the center of the patch were masked out as well.
Final values used were the average directedness calculated for the remaining spatial bins.

\subsection*{Neural representation}

The activation profiles of perceptron nodes were plotted with respect to spatial position in the environment and orientation, 
or head direction as head and body are not separable in our model.
Datasets used for the grid plots (Fig~\ref{fig:neureps}A-C), same as for \ref{fig:traj_vecfield_ori} Fig,
consisted of node outputs for a static, uniform sample of agent locations and orientations.
Datasets for the trajectory plots (Fig~\ref{fig:neureps}D-F) were the same as used for the movement correlations.
Using either dataset, output activity was sorted via 50x50 spatial bins, averaged within each bin, 
smoothed with a Gaussian filter of standard deviation 1, and plotted from zero to maximum activation. 
The data was further sorted into 8 orientation bins and separately plotted. 
This procedure follows a recent paper that focused on constructed activation rate maps from neural networks trained to navigate
\cite{vijayabaskaranNavigationTaskAction2022}.
Neural tuning curves were plotted by binning activation according to the angle between the agent and the goal,
as in \cite{sarelVectorialRepresentationSpatial2017}.

\subsection*{Classification procedure}

Classifying navigational strategies entailed two phases: binning then qualifying.
First, bins were constructed in the 2D phase space of decorrelation time and directedness.
The boundaries were conservatively drawn so as to limit any potential mixing with hybrid strategies.
All agents with decorrelation time below 90 and directedness below 0.6 were labeled as part of the biased diffusive class,
those with decorrelation time between 105 and 140, with directedness above 0.65 were indirect sequential,
and with decorrelation time above 155 and directedness above 0.65 were direct pathing.
Remaining unlabeled runs in the hybrid zones were classified according to a subjective summation of trajectory maps and all movement correlation plots,
as a more quantitative method was deemed to yield marginal relative value.
A small quantity of previously binned and labeled runs were relabeled as hybrids, despite the results of the two metrics.
Relative occurrence plots visualized the same labeled data with respect to visual resolution and distance scaling information.

While the clusters are visually separable, unsupervised clustering algorithms learn to draw in a similar way
(\ref{fig:classes_extended} Fig, bottom row).
Although unsupervised clustering algorithms can designate boundaries, hybrid strategies between or far from cluster centers require qualitative context.

\subsection*{Bottom-up dimensionality reduction}

To complement the classification procedure, nonlinear dimensionality reduction algorithms provide a bottom-up, 
however less explainable means to separate navigational strategy.
The set of actions related to all possible visual observations throughout the environment 
fully describe agent behavior.
Agents with differing visual resolutions have access to differing numbers of possible visual observations.
Lower visual resolution allows a lower total number of views than higher.
A basis was chosen with visual resolution 8 with 132 unique views.
The corresponding mean positions and orientations standardized the set of visual observations for every agent.
The absolute value was taken for the resulting 132 actions, as agents learned to prefer clockwise or counterclockwise turning,
and this break in symmetry is irrelevant to the classification of navigational strategy.

The results of two popular nonlinear dimensionality reduction algorithms, t-SNE and UMAP,
both demonstrate separation aligning with our classification
(\ref{fig:classes_dimred} Fig).
The direct pathing class distinctly separates from the two angle-based classes,
and the indirect sequential and biased diffusive stretch along a gradient in another cluster.
The clear distinction separating distance-based and angle-based navigational strategies,
intuitively aligns with expectations.
The graded cluster of the two angle-based classes suggests a continuum rather than separation,
which may perhaps be more apparent when measures other than decorrelation time and directional entropy are taken into account.
Despite such possibility, both decorrelation time and directional entropy appear to align with the separation and gradient
of the two dimensionality reduction procedures. 
Though the former more smoothly aligns, especially considering the DP/BD hybrid agents. 

Prior to t-SNE or UMAP, the data was passed through PCA to identify the 50 most principal components,
a practice cited to reduce noise and high dimensionality. 
In practice, it did not appear to give a significantly different result.

Gaussian mixtures were applied to the t-SNE and UMAP results, demonstrating that unsupervised classification
draws boundaries in a similar way to our metric-based procedure
(\ref{fig:classes_dimred} Fig).

\section{Acknowledgements}
Ideas within this paper were inspired from discussions at CapoCaccia Cognitive Neuromorphic Workshops (CCNW) 2024,
especially those with Gabriel Gattaux, Andrew Philippides, and Florian Engert.
Discussions with Nereu Montserrat Busquets, Clemence Bergerot, Yunus Sevinchan, Valerii Chirkov, and Valentin Lecheval,
have proven to be invaluable in developing the concept and delivery of the paper.

P.G. acknowledges support by the Elsa-Neumann-Stipendium by the state of Berlin
under Nachwuchsförderungsgesetz (NaFöG) - application number H75014.
P.R. acknowledges funding by the Deutsche Forschungsgemeinschaft (DFG, German Research Foundation)
under Germany’s Excellence Strategy – EXC 2002/1 “Science of Intelligence” – project number 390523135.

All code and data used for this paper is publically available at https://github.com/pgovoni21/vis-nav-abm.

\bibliographystyle{unsrt}
\bibliography{indiv-vis-nav-abm}
% \bibliography{script.bbl}
% \input{script.bbl}

\section{Supporting information}

\setcounter{figure}{0}
\renewcommand{\figurename}{Fig}
\renewcommand{\thefigure}{S\arabic{figure}}

\begin{figure}[tb]
	\centering
	\begin{subfigure}[b]{\linewidth}
	  \centering
	  \includegraphics[width=.33\linewidth,trim={.2cm 0cm 1cm .8cm},clip]{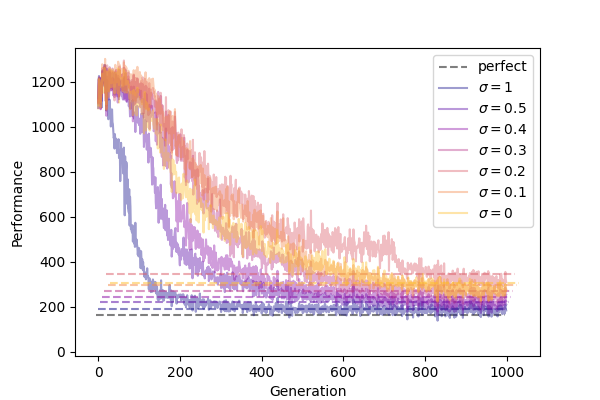}\
	  \includegraphics[width=.33\linewidth,trim={0cm 0cm 1.5cm 1.1cm},clip]{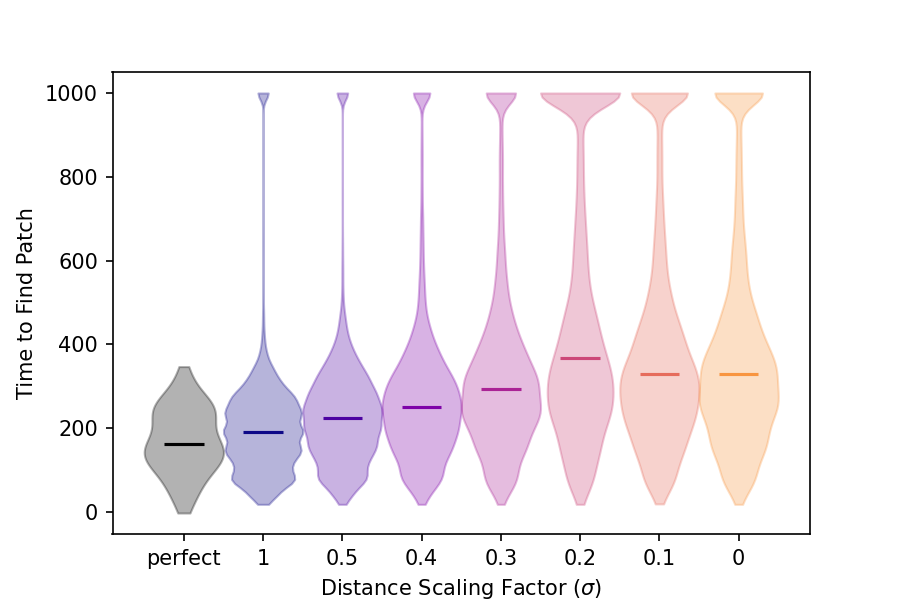}
	\end{subfigure}
	\begin{subfigure}[b]{\linewidth}
	  \centering
	  \includegraphics[width=.33\linewidth,trim={.2cm 0cm 1cm .8cm},clip]{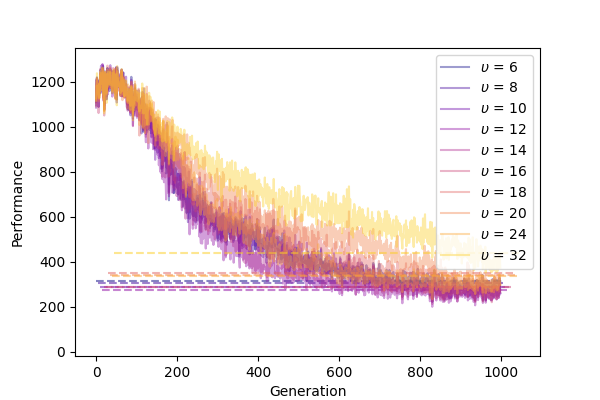}
	  \includegraphics[width=.33\linewidth,trim={0cm 0cm 1.5cm .8cm},clip]{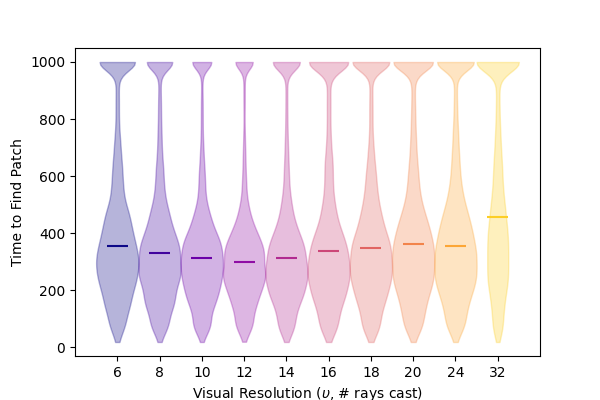}
	\end{subfigure}
	\begin{subfigure}[b]{\linewidth}
	  \centering
	  \includegraphics[width=.33\linewidth,trim={.2cm 0cm 1cm 1.1cm},clip]{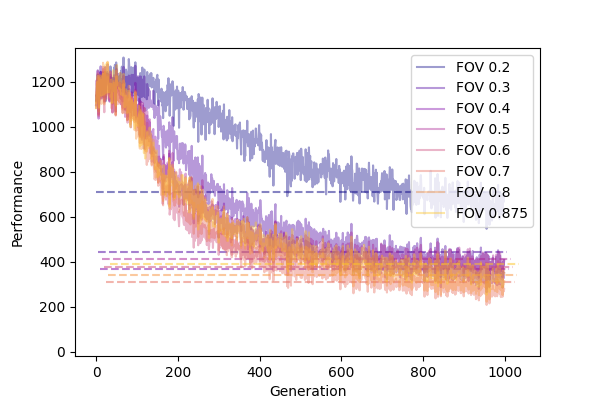}
	  \includegraphics[width=.33\linewidth,trim={0cm 0cm 1.5cm 1.1cm},clip]{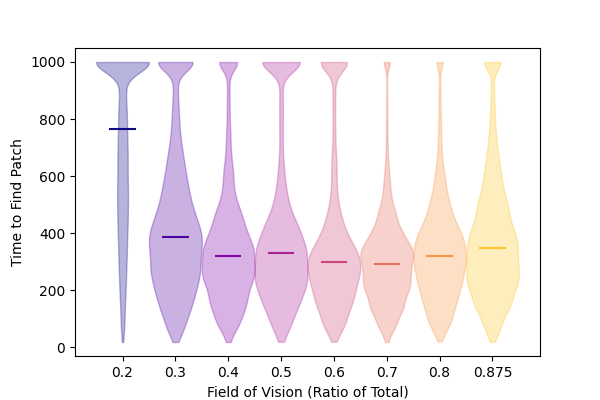}
	\end{subfigure}
	\begin{subfigure}[b]{\linewidth}
	  \centering
	  \includegraphics[width=.33\linewidth,trim={.2cm 0cm 1cm .8cm},clip]{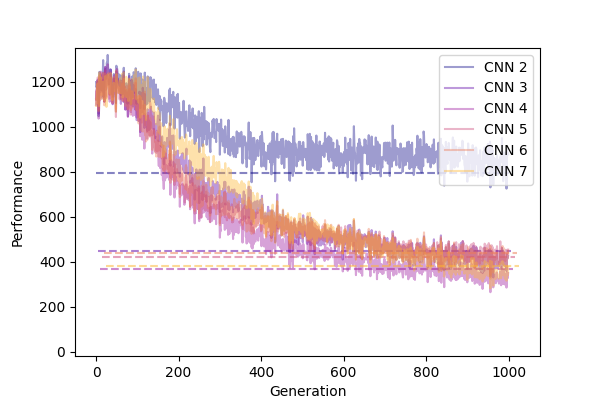}
	  \includegraphics[width=.33\linewidth,trim={0cm 0cm 1.5cm .8cm},clip]{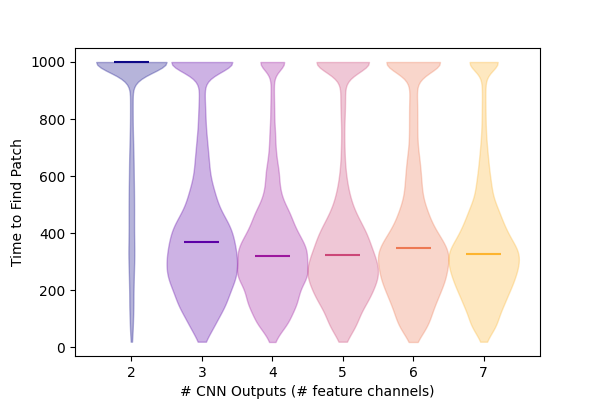}
	\end{subfigure}
	\begin{subfigure}[b]{\linewidth}
	  \centering
	  \includegraphics[width=.33\linewidth,trim={.2cm 0cm 1cm .8cm},clip]{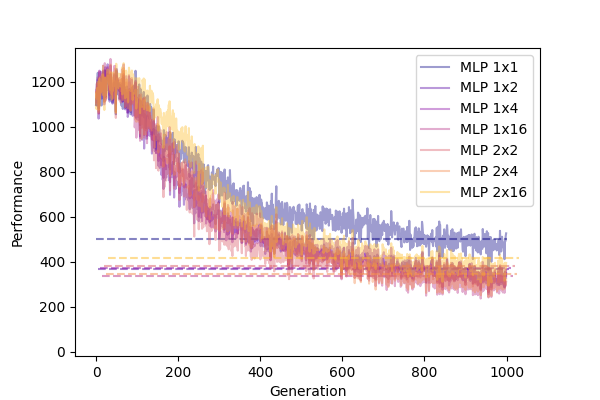}
	  \includegraphics[width=.33\linewidth,trim={0cm 0cm 1.5cm .8cm},clip]{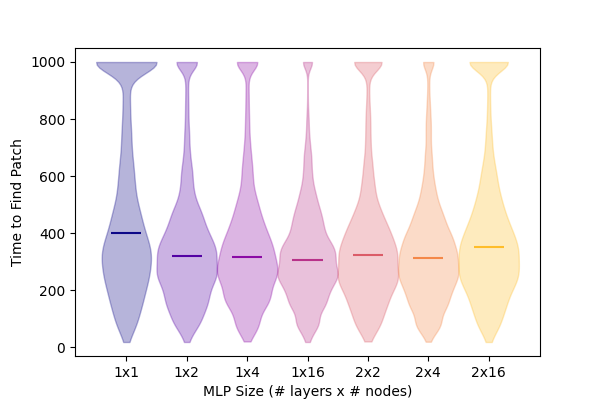}
	\end{subfigure}
	
	\caption{
		\textbf{Evolutionary performance with differing parameters.}
		Left: median performance of 40 training runs, dashed lines indicate median of validation tests.
		Right: validation test distribution (remaining distance to patch not included), lines indicate median, same color as legend in left plot.
		Runs used in top figure reflect data used in main text (perfect: theoretical lower bound).
		Unless otherwise noted, parameters are as follows:
		Distance scaling factor ($\sigma$): 0; 
		Visual resolution ($\upsilon$): 8; 
		Field of vision: 0.4; 
		CNN output size: 4;
		Multilayer perceptron (MLP) size: 1x2.
		}
	\label{fig:evol_params}
  \end{figure}

  \begin{figure}[tb]
	\centering
	\begin{subfigure}[b]{\linewidth}
	  \centering
	  \includegraphics[width=0.3\linewidth,trim={90 77 70 80},clip]{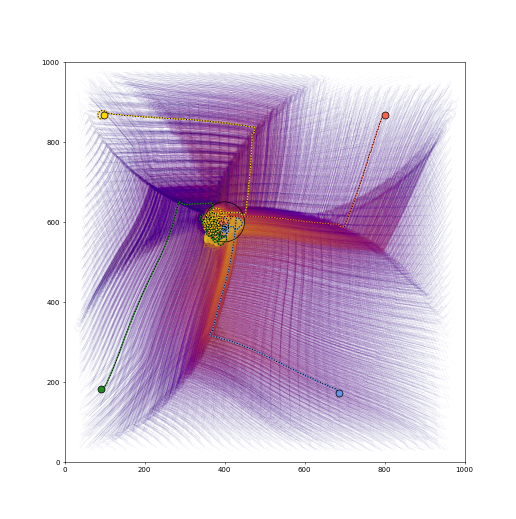}
	  \includegraphics[width=0.3\linewidth,trim={90 77 70 80},clip]{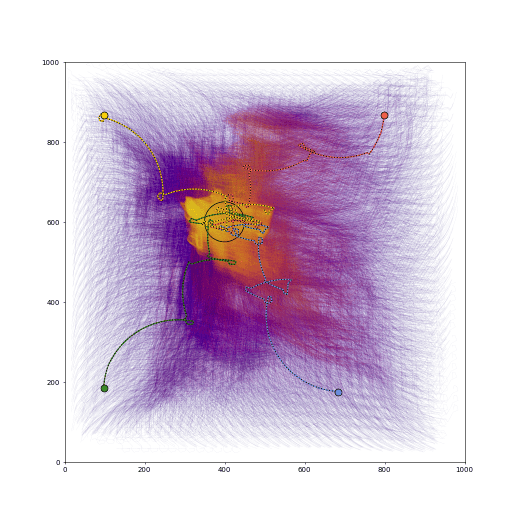}
	  \includegraphics[width=0.3\linewidth,trim={90 77 70 80},clip]{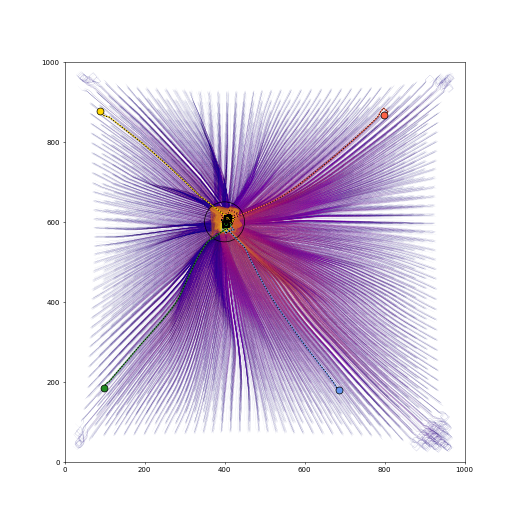}
	\end{subfigure}
  
	\begin{subfigure}[b]{\linewidth}
	  \centering
	  \includegraphics[width=0.3\linewidth,trim={90 77 70 80},clip]{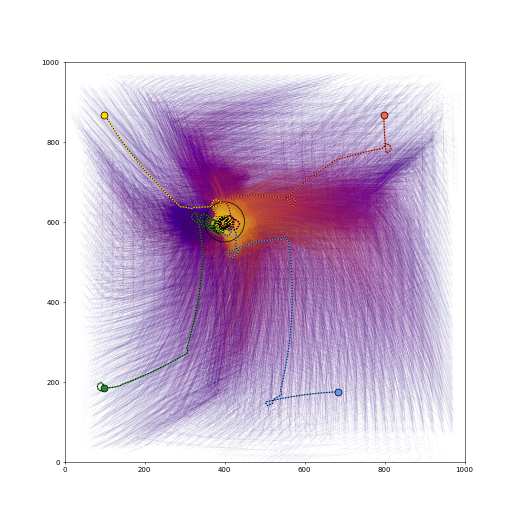}
	  \includegraphics[width=0.3\linewidth,trim={90 77 70 80},clip]{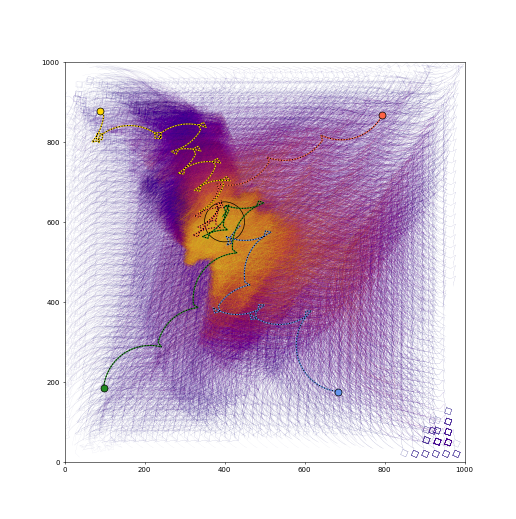}
	  \includegraphics[width=0.3\linewidth,trim={90 77 70 80},clip]{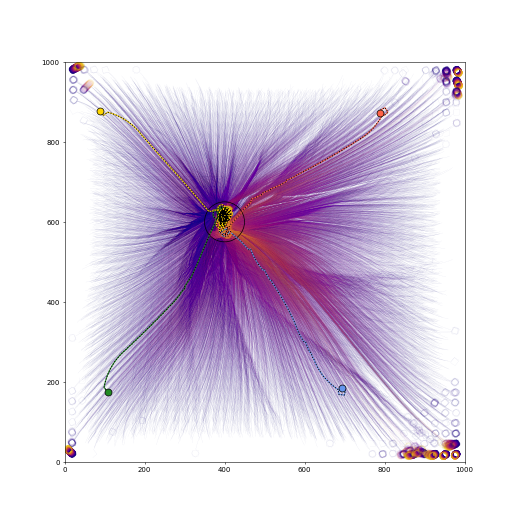}
	\end{subfigure}
  
	\begin{subfigure}[b]{0.3\linewidth}
	  \includegraphics[width=\linewidth,trim={90 77 70 80},clip]{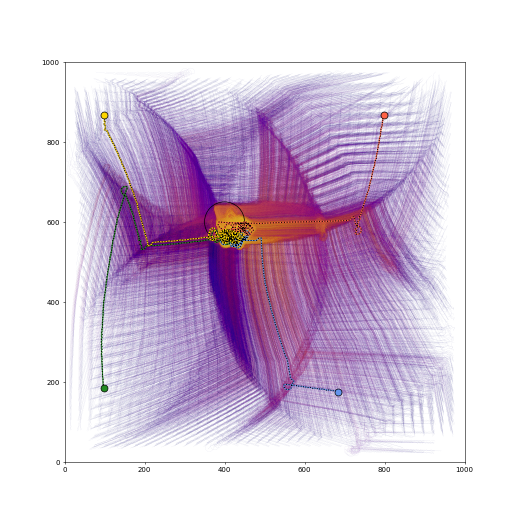}
	  \caption{Indirect Sequential}
	\end{subfigure}
	\begin{subfigure}[b]{0.3\linewidth}
	  \includegraphics[width=\linewidth,trim={90 77 70 80},clip]{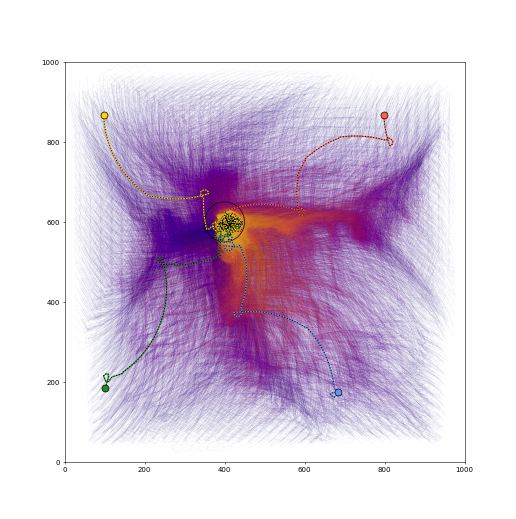}
	  \caption{Biased Diffusive}
	\end{subfigure}
	\begin{subfigure}[b]{0.3\linewidth}
	  \includegraphics[width=\linewidth,trim={90 77 70 80},clip]{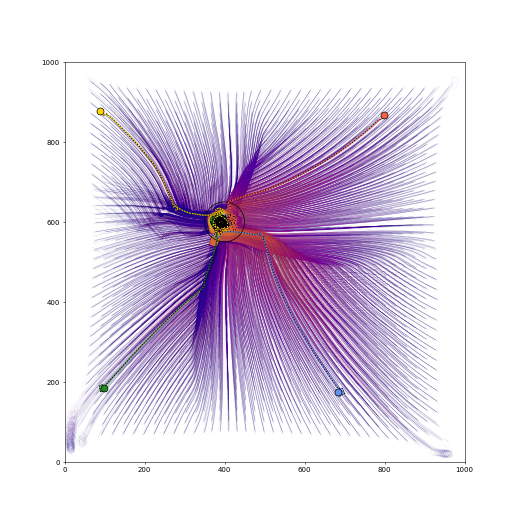}
	  \caption{Direct Pathing}
	\end{subfigure}

	\caption{
		\textbf{Extended figure from Fig \ref{fig:trajs_corrs} top row trajectories.}
		Top/middle rows: $\upsilon= 8$. 
		Bottom row: $\upsilon= 16$.
		See \ref{fig:corrs_ext} Fig for corresponding correlations.
		}
	\label{fig:trajs_ext}
  \end{figure}

  \begin{figure}[htb]
	\centering
	\includegraphics[width=\linewidth,trim={0 0 0 0},clip]{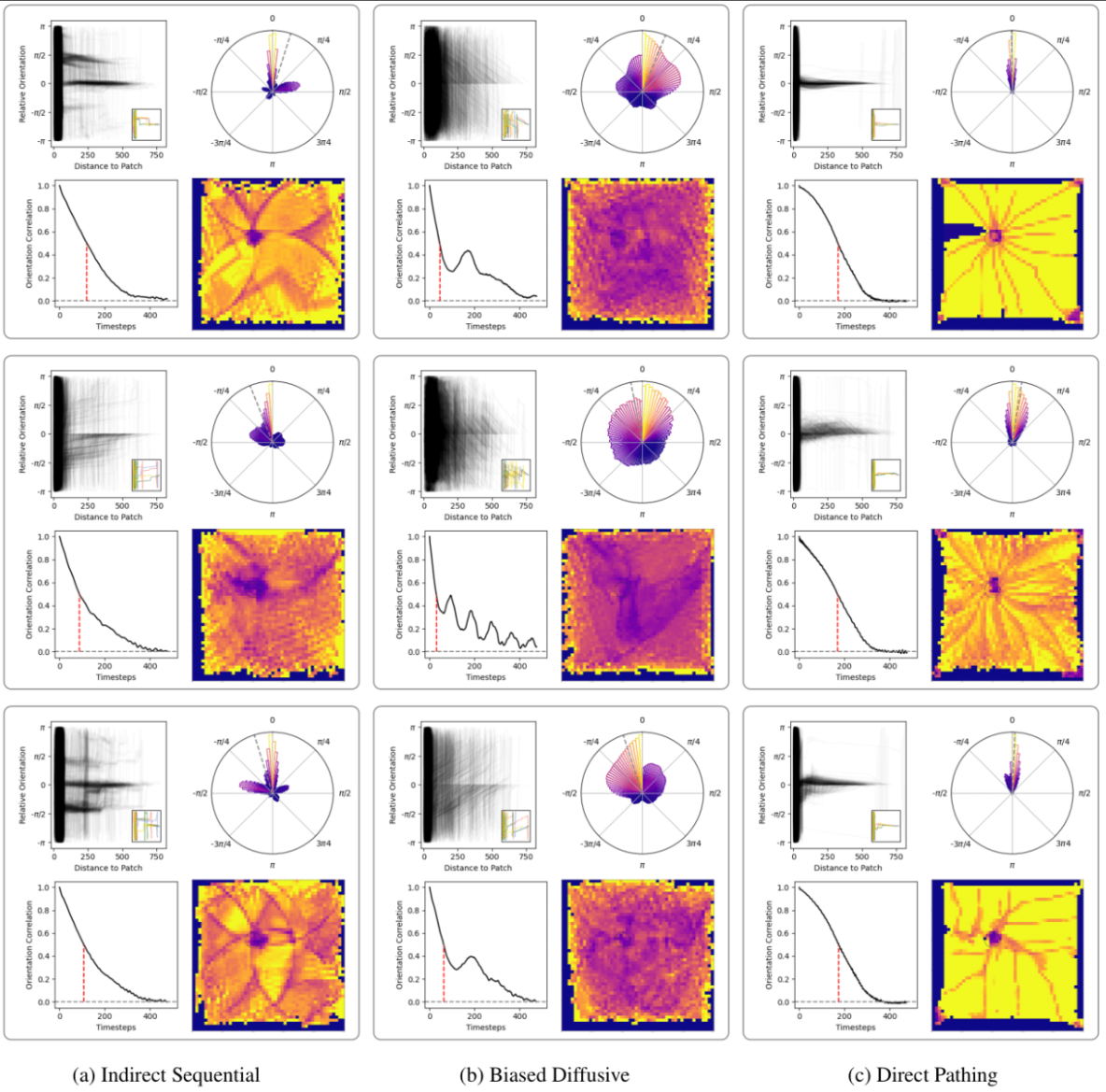}
	\caption{
		\textbf{Extended figure from Fig \ref{fig:trajs_corrs} bottom row correlations.}
		Top/middle rows: $\upsilon= 8$. 
		Bottom row: $\upsilon= 16$.
		See \ref{fig:trajs_ext} Fig for corresponding trajectories.
		}
	\label{fig:corrs_ext}
  \end{figure}

  \begin{figure}[htb]
	\centering
	\includegraphics[width=\linewidth,trim={0 0 0 0},clip]{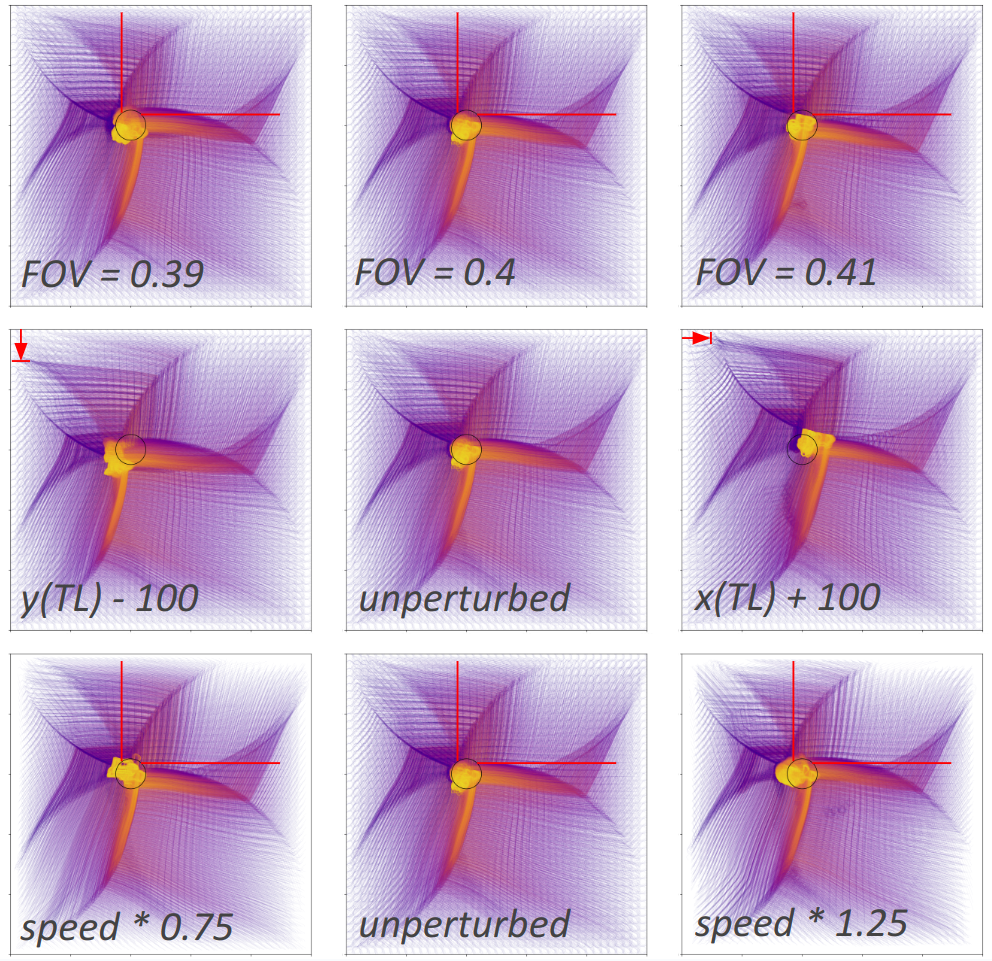}
	\caption{
		\textbf{Perturbations to a trained agent (perception, environment, action).}
		Top row: perturbation of field of vision.
		Middle row: perturbation of NW (top left) corner.
		Bottom row: perturbation of agent speed.
		Middle column: unperturbed agent.
		Red lines: visual guides.
		}
	\label{fig:perturbs}
  \end{figure}

  \begin{figure}[htb]
	\centering
	\includegraphics[width=0.95\linewidth,trim={0 0 0 0},clip]{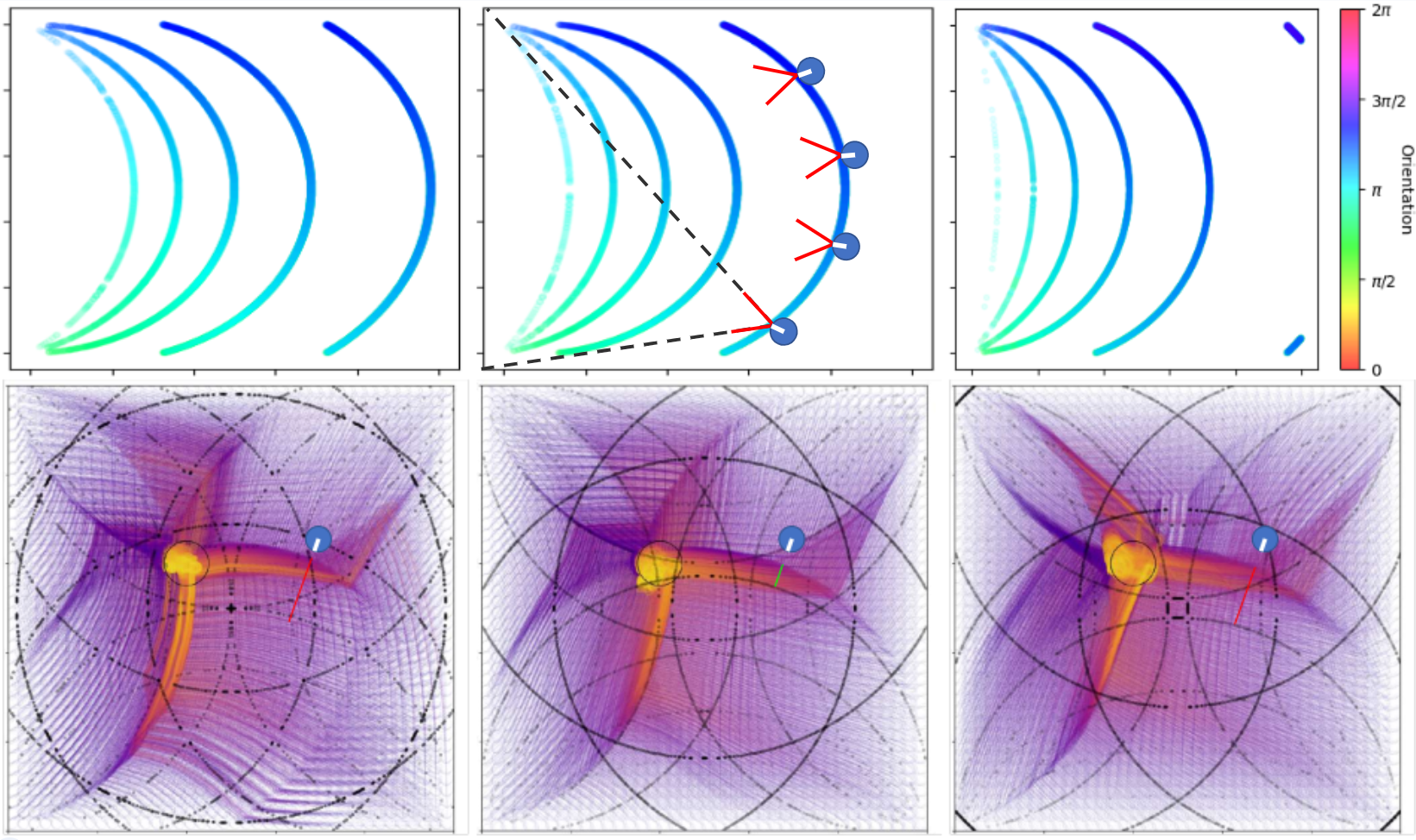}
	\includegraphics[width=0.95\linewidth,trim={0 0 0 0},clip]{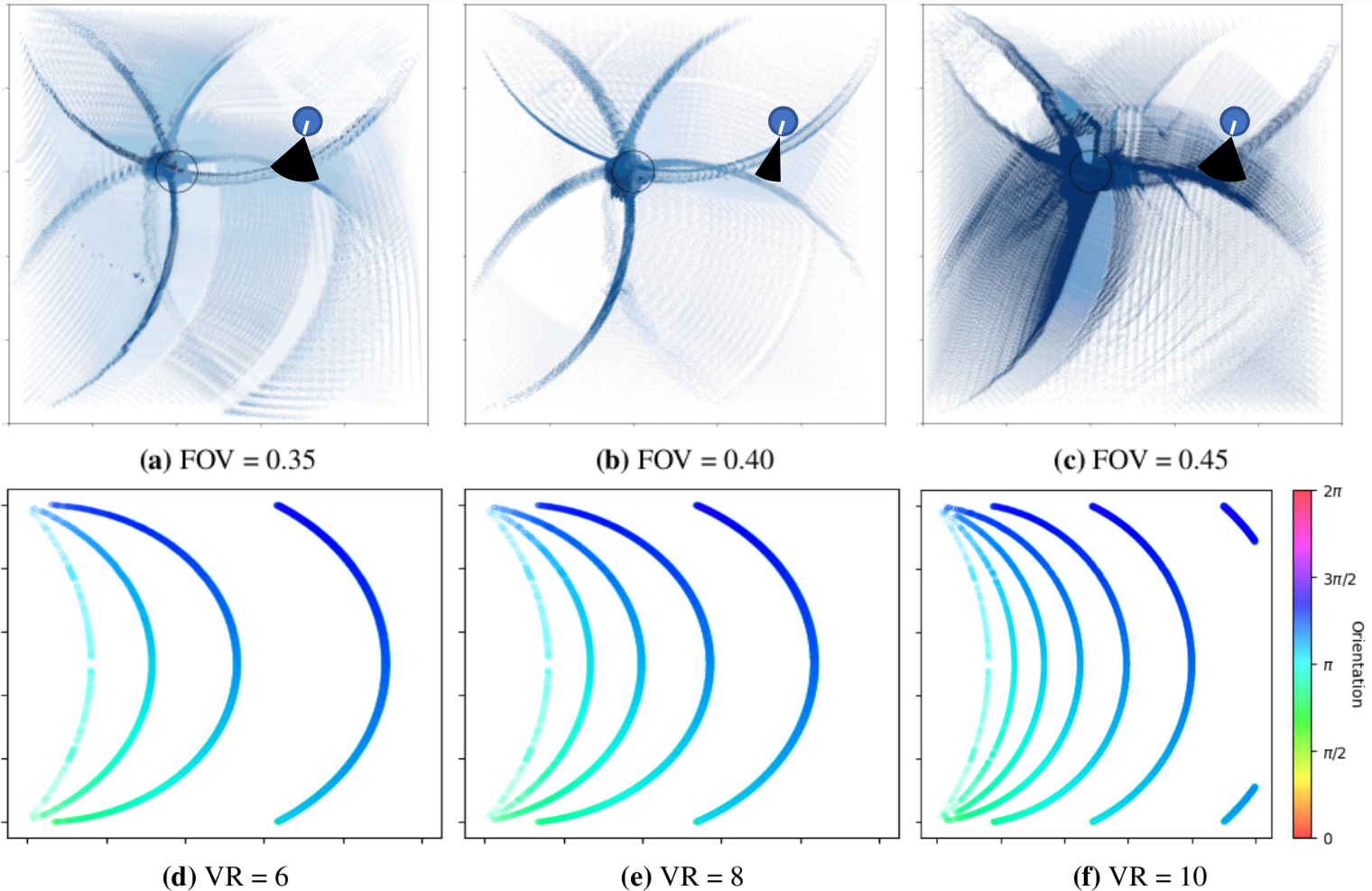}
	\caption{
		\textbf{Dual corner raycast as elliptical decision manifold.}
		1st row: ideal locations for double detection of the top-left/bottom-left corners, for differing field of vision parameters,
		where the agents in (b) illustrate example agent positions and orientations along one of the available elliptical arcs.
		As the agent is rotated, visual observation remains the same, allowing a consistent action to be tied across the manifold.
		2nd row: trajectory maps with points plotted for simultaneous dual corner raycasts. 
		3rd row: absolute value of turning speed, normalized between 0 and 90 degrees per timestep.
		4th row: same as 1st but for differing visual resolutions.
		}
	\label{fig:map_rays}
  \end{figure}

  \begin{figure}[tb]
	\centering
	\begin{subfigure}[b]{\linewidth}
	  \centering
	  \includegraphics[width=0.28\linewidth,trim={55 60 70 70},clip]{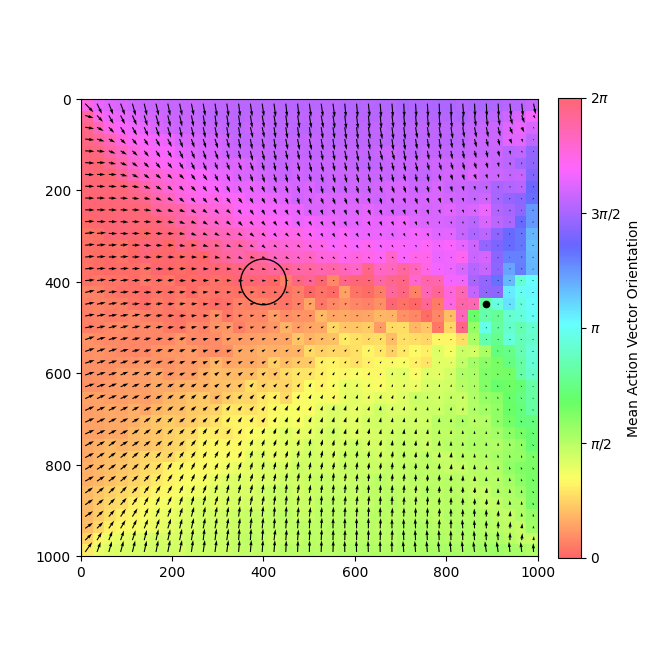}
	  \includegraphics[width=0.28\linewidth,trim={55 60 70 70},clip]{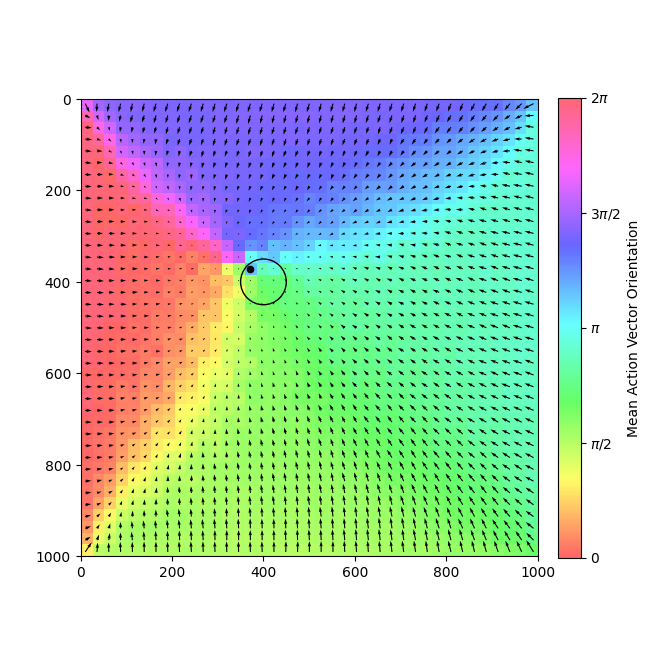}
	  \includegraphics[width=0.28\linewidth,trim={55 60 70 70},clip]{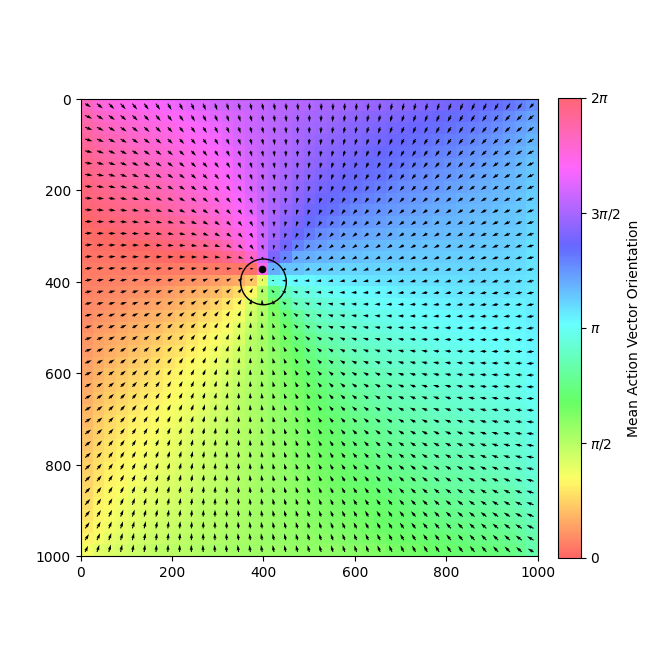}
	\end{subfigure}
  
	\begin{subfigure}[b]{\linewidth}
	  \centering
	  \includegraphics[width=0.28\linewidth,trim={55 60 70 70},clip]{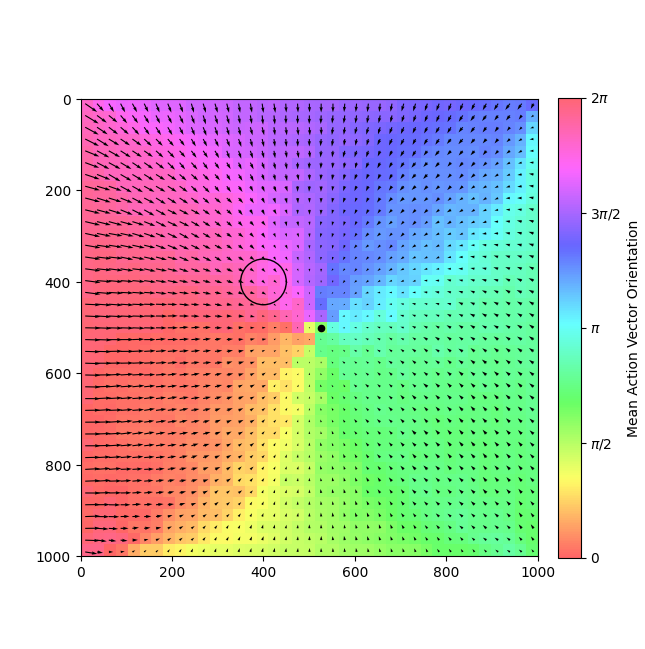}
	  \includegraphics[width=0.28\linewidth,trim={55 60 70 70},clip]{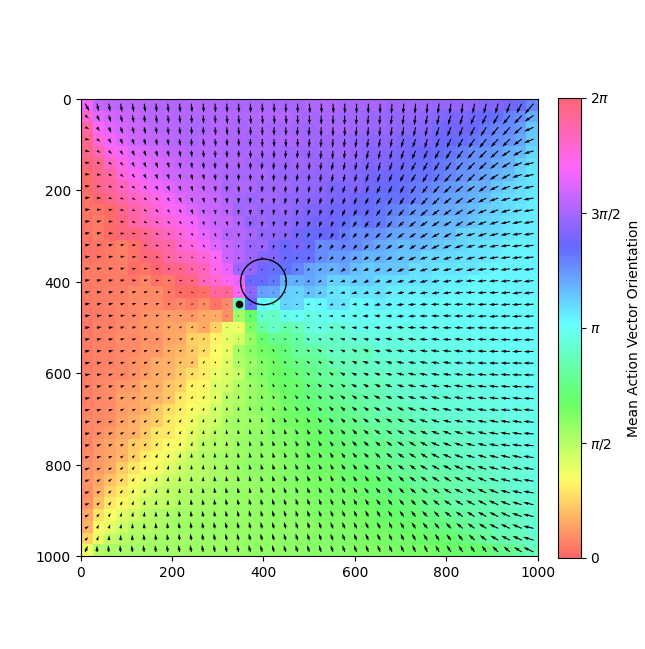}
	  \includegraphics[width=0.28\linewidth,trim={55 60 70 70},clip]{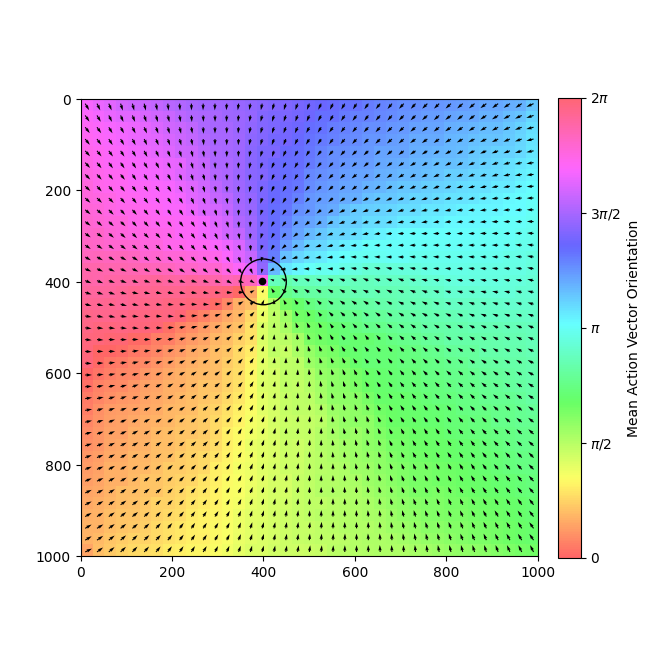}
	\end{subfigure}
  
	\begin{subfigure}[b]{\linewidth}
	  \centering
	  \includegraphics[width=0.28\linewidth,trim={55 60 70 70},clip]{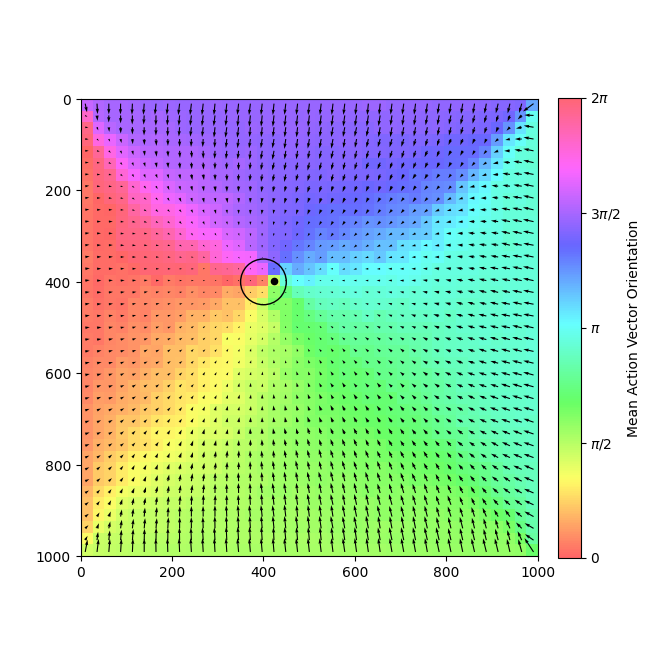}
	  \includegraphics[width=0.28\linewidth,trim={55 60 70 70},clip]{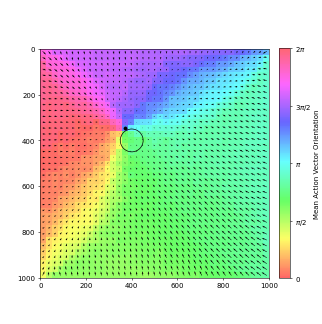}
	  \includegraphics[width=0.28\linewidth,trim={55 60 70 70},clip]{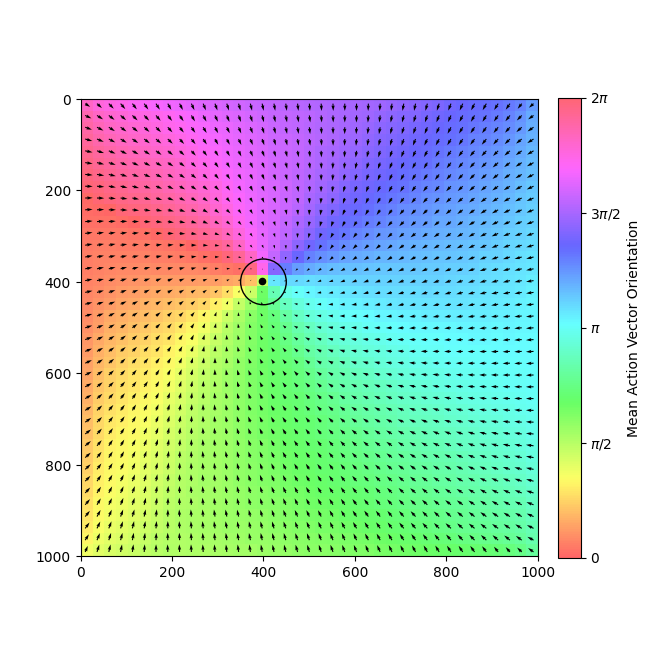}
	\end{subfigure}
  
	\begin{subfigure}[b]{0.28\linewidth}
	  \includegraphics[width=\linewidth,trim={55 60 70 70},clip]{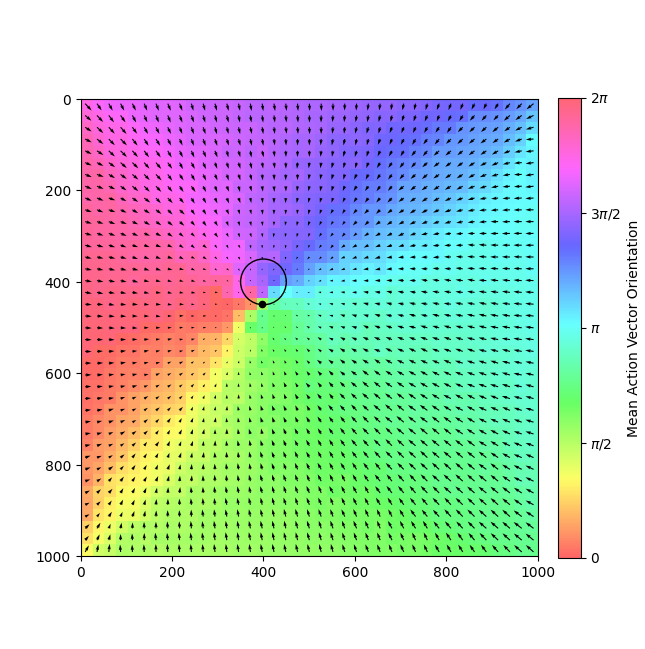}
	  \caption{Indirect Sequential}
	\end{subfigure}
	\begin{subfigure}[b]{0.28\linewidth}
	  \includegraphics[width=\linewidth,trim={55 60 70 70},clip]{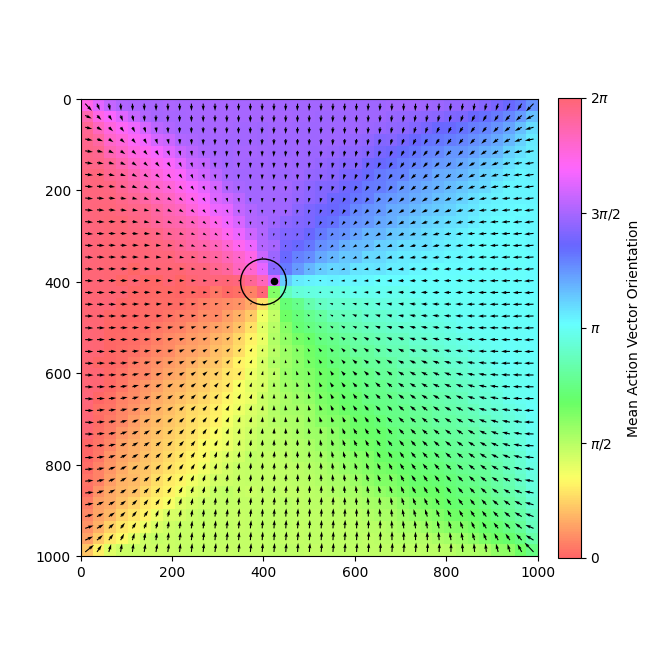}
	  \caption{Biased Diffusive}
	\end{subfigure}
	\begin{subfigure}[b]{0.28\linewidth}
	  \includegraphics[width=\linewidth,trim={55 60 70 70},clip]{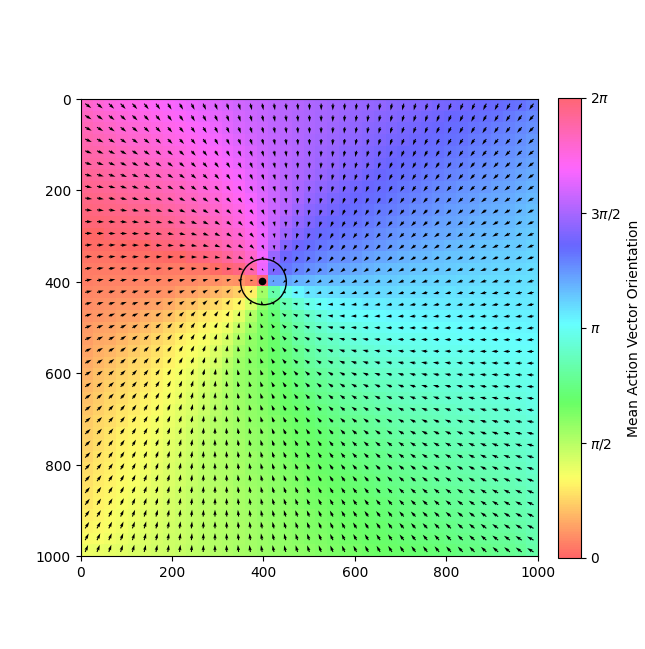}
	  \caption{Direct Pathing}
	\end{subfigure}

	\caption{
		\textbf{Mean action vector fields.}
		Mean vectors are computed and plotted for agent action outputs at 64 unique orientations for each grid location.
		Mean vector orientation are plotted with colors following the colorbar in Fig \ref{fig:map_rays}.
		Agents used mirror those in Fig \ref{fig:trajs_corrs} (top row) \& \ref{fig:trajs_ext} Fig (bottom three).
		Black points signify global minima in mean vector magnitude.
		}
	\label{fig:traj_vecfield_ori}
  \end{figure}

  \begin{figure}[t]
	\centering
	\begin{subfigure}[b]{\linewidth}
	  \centering
	  \includegraphics[width=0.3\linewidth,trim={0 10 0 10},clip]{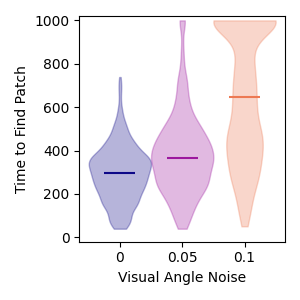}
	  \includegraphics[width=0.3\linewidth,trim={0 10 0 10},clip]{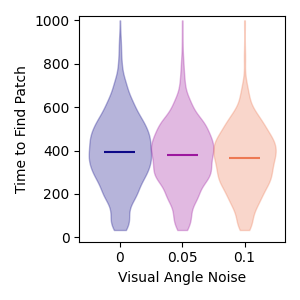}
	  \includegraphics[width=0.3\linewidth,trim={0 10 0 10},clip]{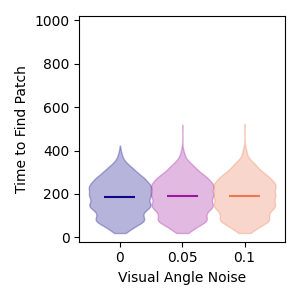}
	\end{subfigure}
  
	\begin{subfigure}[b]{\linewidth}
	  \centering
	  \includegraphics[width=0.3\linewidth,trim={0 10 0 10},clip]{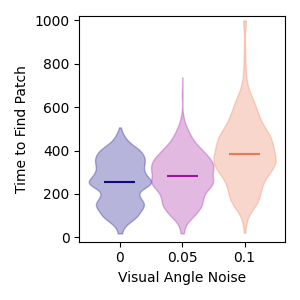}
	  \includegraphics[width=0.3\linewidth,trim={0 10 0 10},clip]{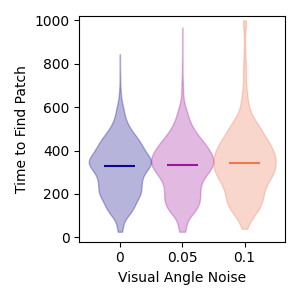}
	  \includegraphics[width=0.3\linewidth,trim={0 10 0 10},clip]{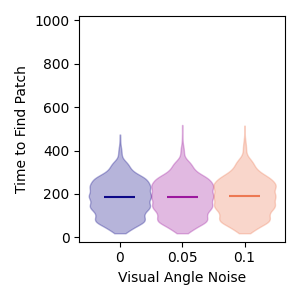}
	\end{subfigure}
  
	\begin{subfigure}[b]{\linewidth}
	  \centering
	  \includegraphics[width=0.3\linewidth,trim={0 10 0 0},clip]{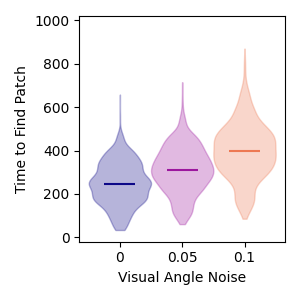}
	  \includegraphics[width=0.3\linewidth,trim={0 10 0 0},clip]{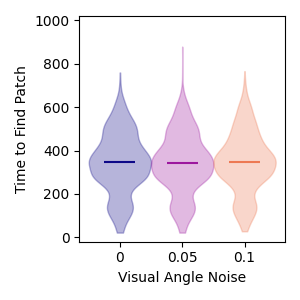}
	  \includegraphics[width=0.3\linewidth,trim={0 10 0 0},clip]{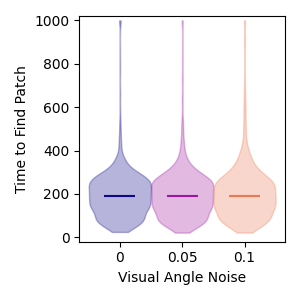}
	\end{subfigure}
  
	\begin{subfigure}[b]{0.3\linewidth}
	  \includegraphics[width=\linewidth,trim={0 10 0 0},clip]{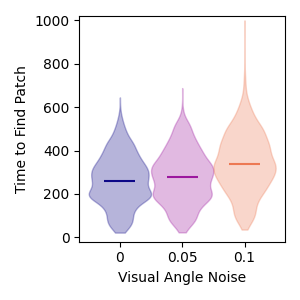}
	  \caption{Indirect Sequential}
	\end{subfigure}
	\begin{subfigure}[b]{0.3\linewidth}
	  \includegraphics[width=\linewidth,trim={0 10 0 0},clip]{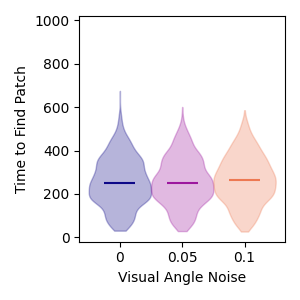}
	  \caption{Biased Diffusive}
	\end{subfigure}
	\begin{subfigure}[b]{0.3\linewidth}
	  \includegraphics[width=\linewidth,trim={0 10 0 0},clip]{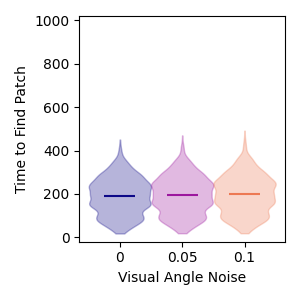}
	  \caption{Direct Pathing}
	\end{subfigure}
	
	\caption{
		\textbf{Effect of noise to the visual angle.}
		Visual angle, as the median of an agent FOV, is perturbed by Gaussian noise with the two standard deviation of 0.05 and 0.1 radians.
		Same set of initializations/networks as in Fig \ref{fig:trajs_corrs} (top row) \& \ref{fig:trajs_ext} Fig (bottom three).
		}
	\label{fig:noise_visangle}
  \end{figure}

  \begin{figure}[htb]
	\centering
	\begin{subfigure}[b]{\linewidth}
	  \centering
	  \includegraphics[width=0.24\linewidth,trim={3.1cm 2.7cm 2cm 2.5cm},clip]{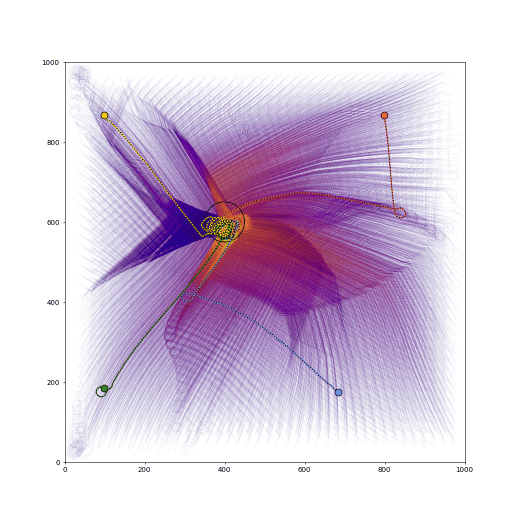}
	  \includegraphics[width=0.24\linewidth,trim={3.1cm 2.7cm 2cm 2.5cm},clip]{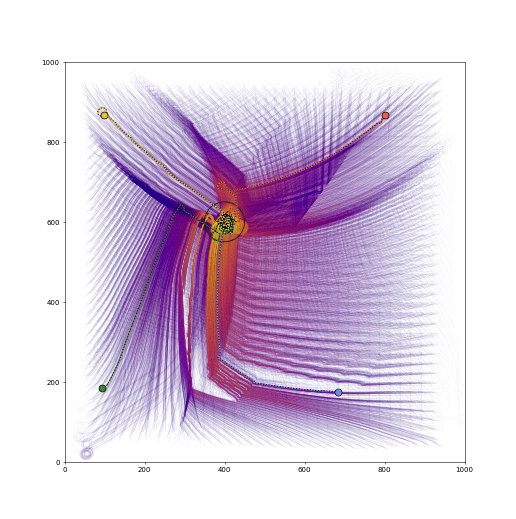}
	  \includegraphics[width=0.24\linewidth,trim={3.1cm 2.7cm 2cm 2.5cm},clip]{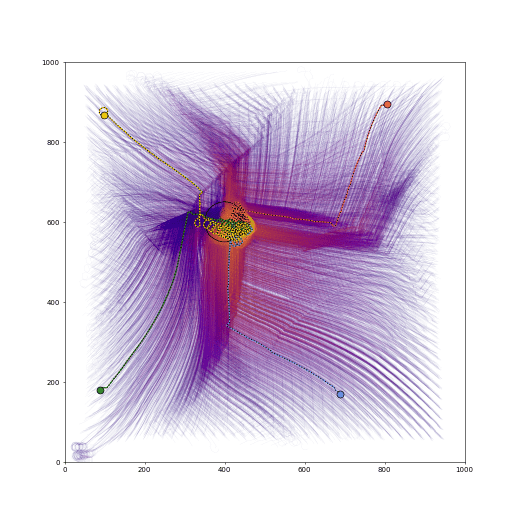}
	  \includegraphics[width=0.24\linewidth,trim={2.2cm 1.9cm 1.3cm 1.5cm},clip]{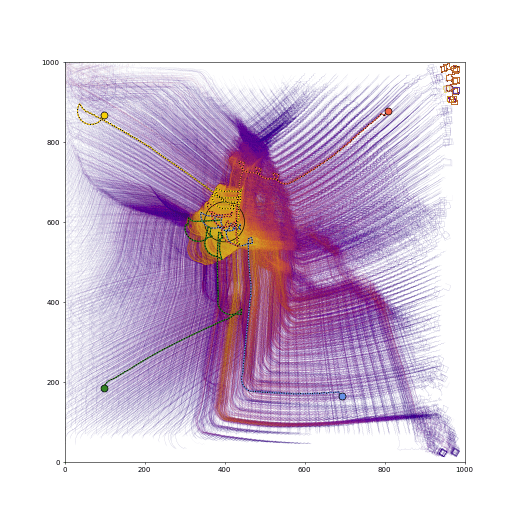}
	\end{subfigure}
	\begin{subfigure}[b]{\linewidth}
	  \centering
	  \includegraphics[width=0.24\linewidth,trim={3.1cm 2.7cm 2cm 2.5cm},clip]{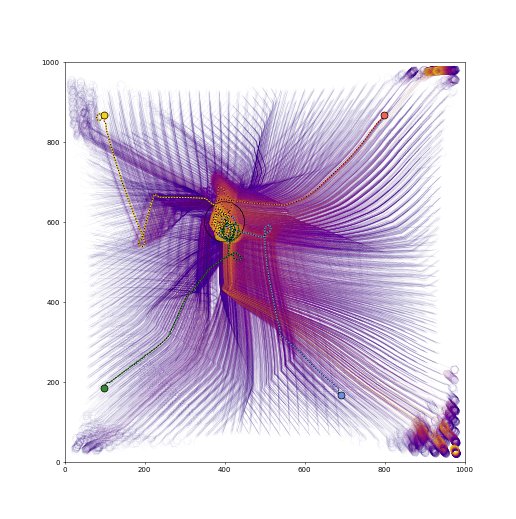}
	  \includegraphics[width=0.24\linewidth,trim={3.1cm 2.7cm 2cm 2.5cm},clip]{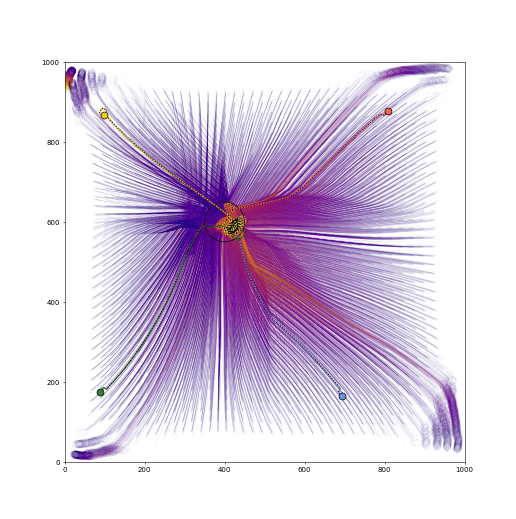}
	  \includegraphics[width=0.24\linewidth,trim={3.1cm 2.7cm 2cm 2.5cm},clip]{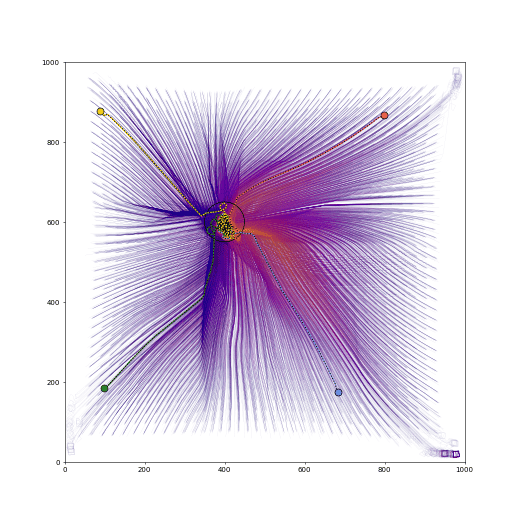}
	  \includegraphics[width=0.24\linewidth,trim={3.1cm 2.7cm 2cm 2.5cm},clip]{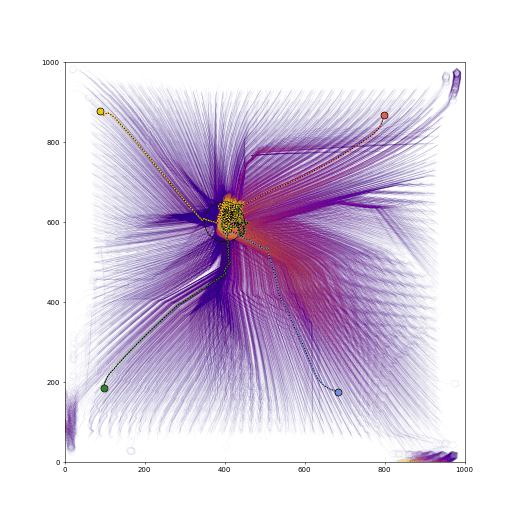}
	\end{subfigure}
	\begin{subfigure}[b]{\linewidth}
	  \centering
	  \includegraphics[width=0.24\linewidth,trim={3.1cm 2.7cm 2cm 2.5cm},clip]{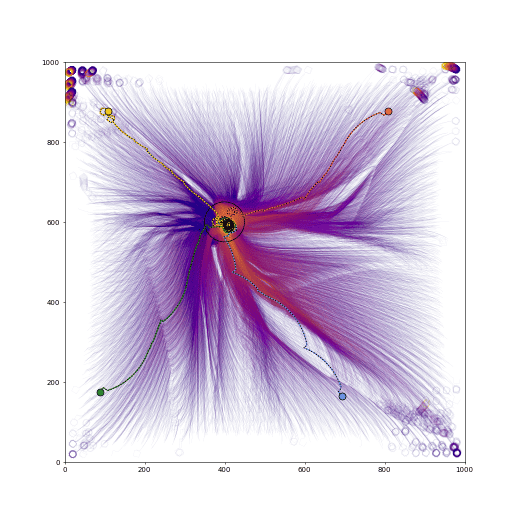}
	  \includegraphics[width=0.24\linewidth,trim={2.2cm 1.9cm 1.4cm 2cm},clip]{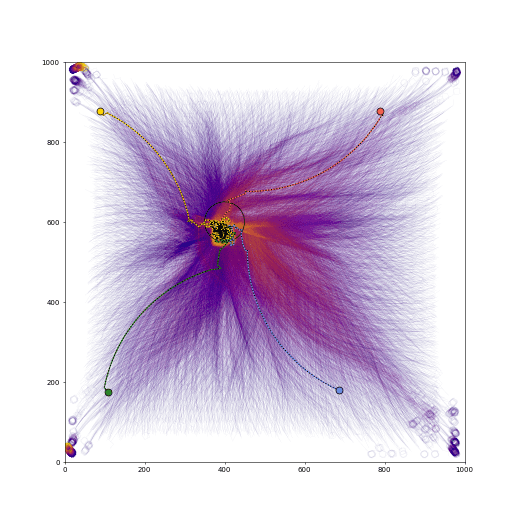}
	  \includegraphics[width=0.24\linewidth,trim={3.1cm 2.7cm 2cm 2.5cm},clip]{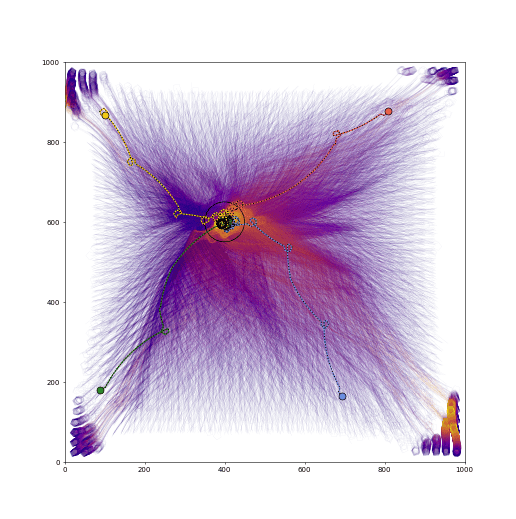}
	  \includegraphics[width=0.24\linewidth,trim={3.1cm 2.7cm 2cm 2.5cm},clip]{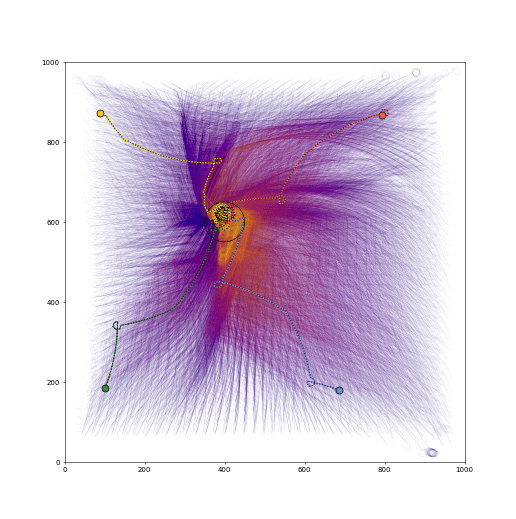}
	\end{subfigure}
	\caption{
		\textbf{A collection of trajectory plots representing behavioral hybridization possible.}
		$\sigma$ damped to 0.5, midway between Fig \ref{fig:trajs_corrs} A/B \& C, at visual resolution 8,
		arranged as a double continuum of classes: from indirect sequential to direct pathing to biased diffusion.
		}
	\label{fig:maps_IS_DP_BD}
  \end{figure}

  \begin{figure}[t]
	\centering
	\begin{subfigure}[b]{.49\linewidth}
	  \includegraphics[width=\linewidth,trim={35 10 50 30},clip]{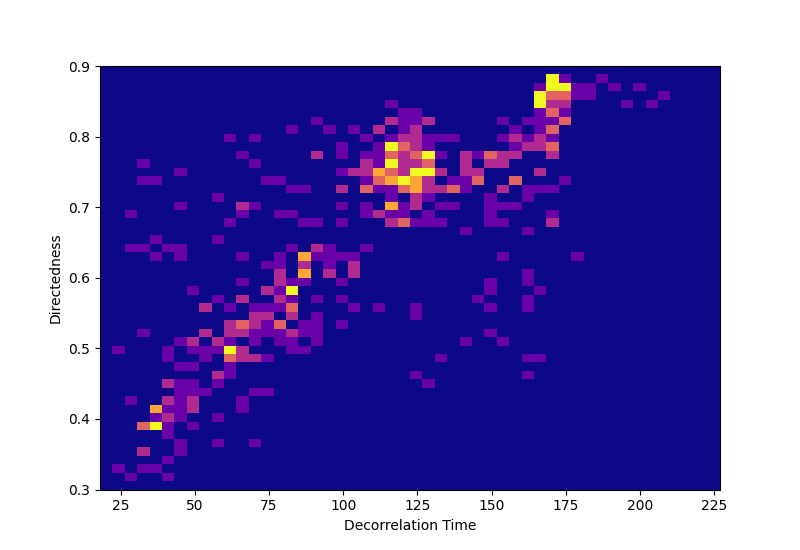}
	\end{subfigure}
	\begin{subfigure}[b]{.49\linewidth}
	  \includegraphics[width=\linewidth,trim={35 10 50 30},clip]{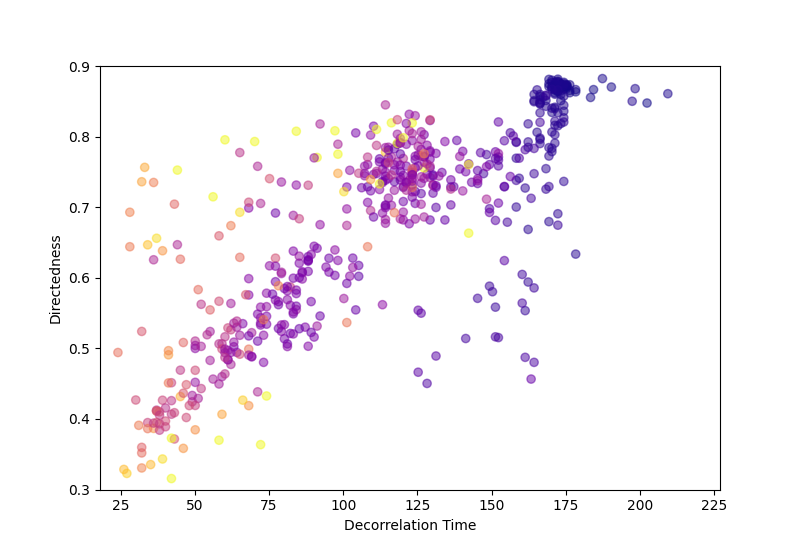}
	\end{subfigure}
	\begin{subfigure}[b]{.49\linewidth}
	  \includegraphics[width=\linewidth,trim={35 10 50 30},clip]{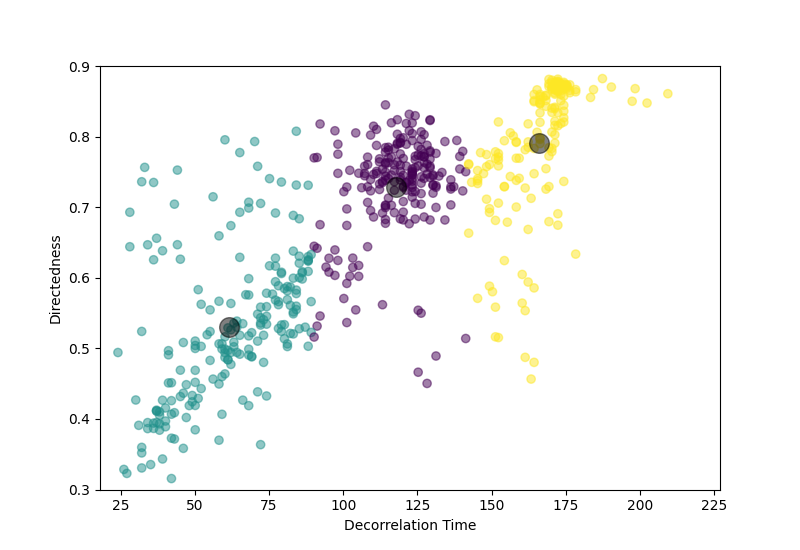}
	\end{subfigure}
	\begin{subfigure}[b]{.49\linewidth}
	  \includegraphics[width=\linewidth,trim={35 10 50 30},clip]{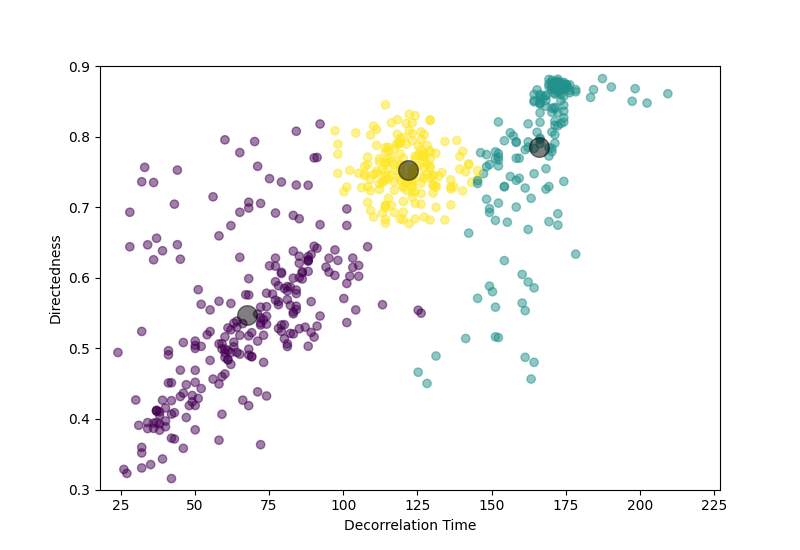}
	\end{subfigure}
	
	\caption{
		\textbf{Extended data for Fig \ref{fig:classes}.}
		Top left: density-based heatmap, maximum in yellow normalized at 5.
		Top right: relative fitnesses (blue: maximum, yellow: minimum).
		Bottom row: unsupervised clustering (left: K-Means, right: gaussian mixture), cluster centers plotted as larger circles.
		}
	\label{fig:classes_extended}
  \end{figure}

\begin{figure}[tb]
	\centering
	\begin{subfigure}[b]{.7\linewidth}
	  \centering
	  \includegraphics[width=\linewidth,trim={17.1cm 0cm 2.5cm 1.2cm},clip]{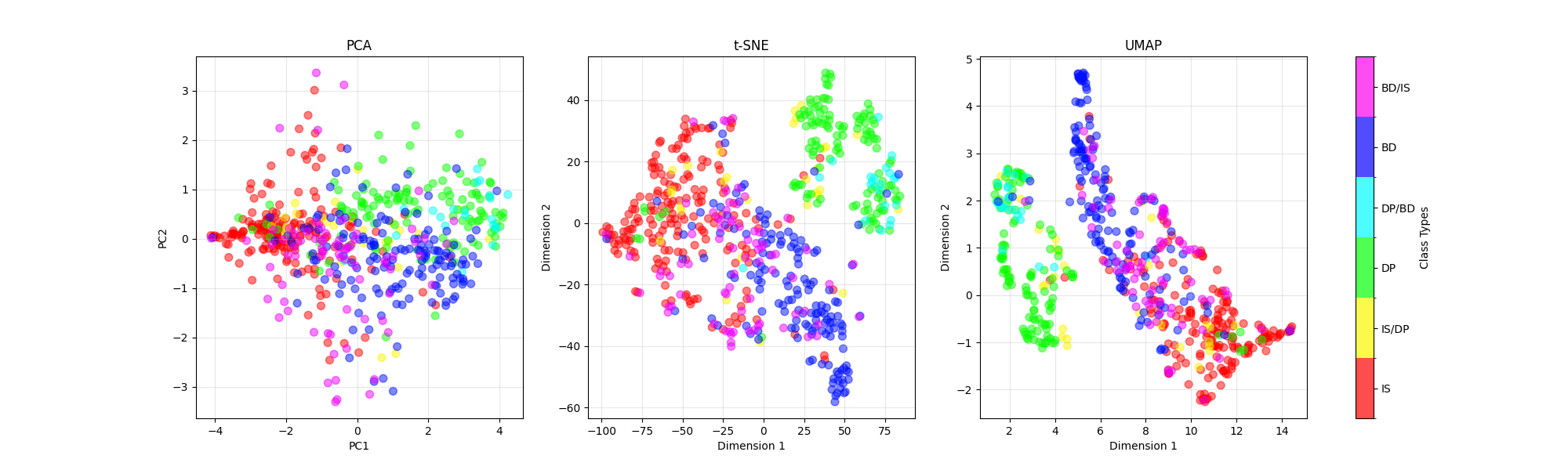}
	\end{subfigure}
	\begin{subfigure}[b]{.7\linewidth}
	  \centering
	  \includegraphics[width=\linewidth,trim={17.1cm 0cm 2.5cm 1.8cm},clip]{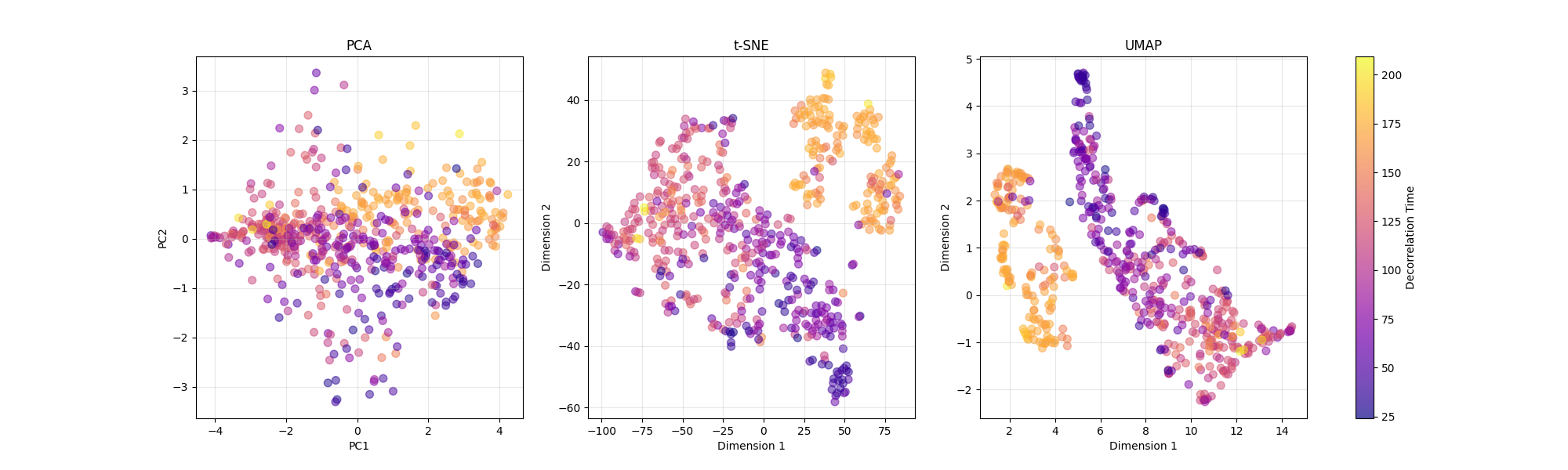}
	\end{subfigure}
	\begin{subfigure}[b]{.7\linewidth}
	  \centering
	  \includegraphics[width=\linewidth,trim={17.1cm 0cm 2.5cm 1.8cm},clip]{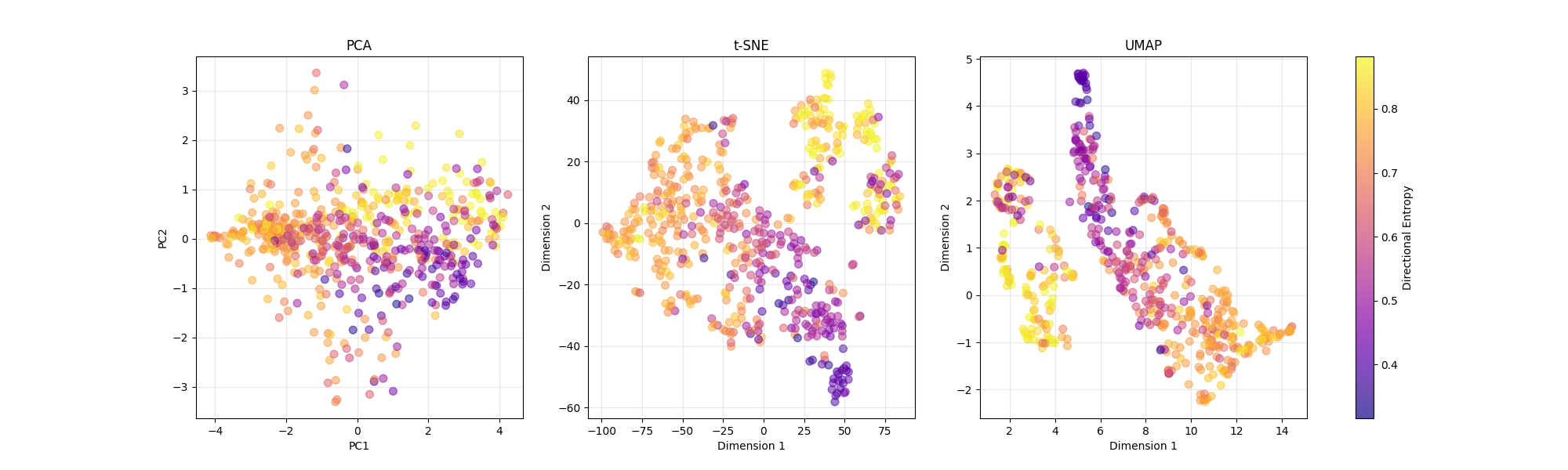}
	\end{subfigure}
	\begin{subfigure}[b]{.7\linewidth}
	  \centering
	  \includegraphics[width=\linewidth,trim={18.5cm 0cm 0cm 15.5cm},clip]{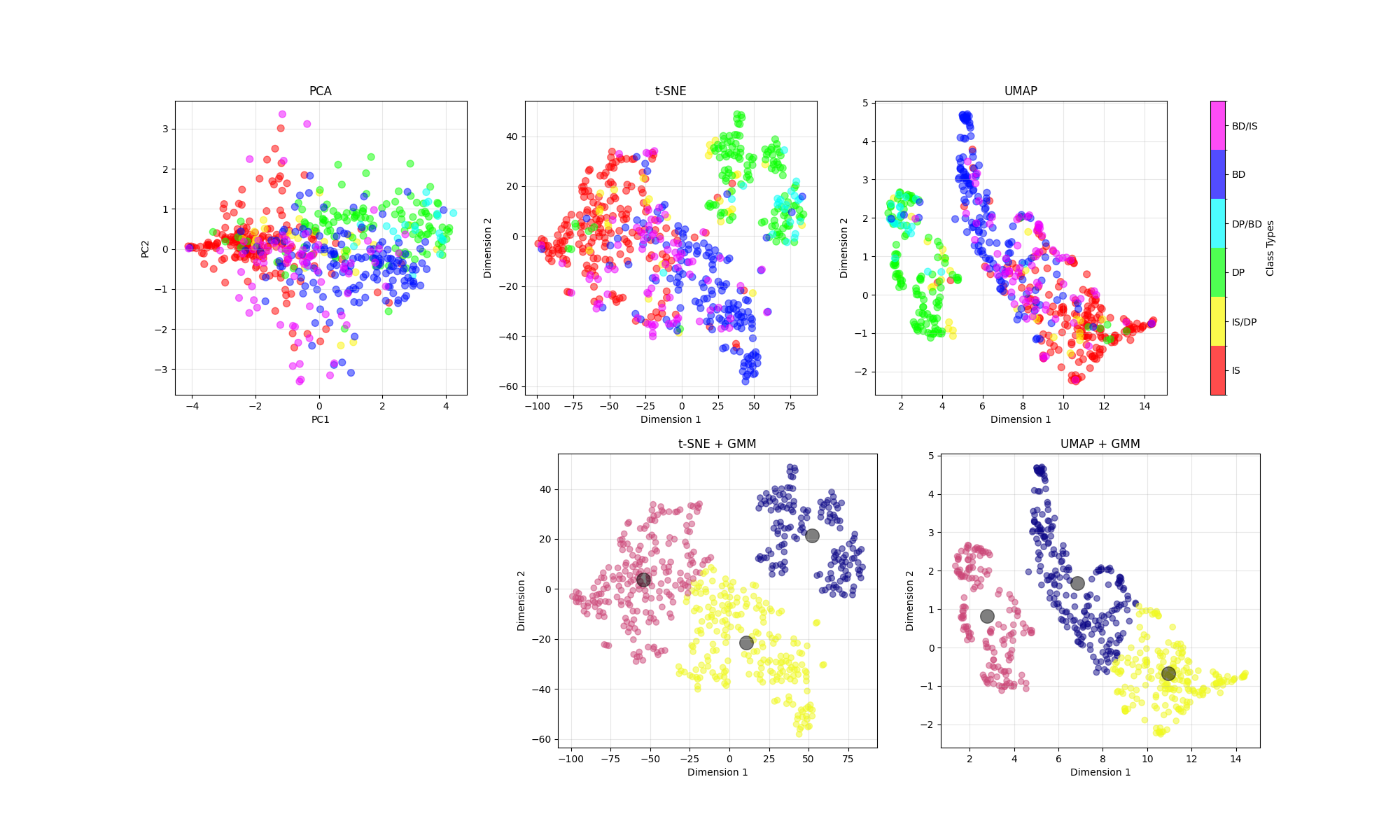}
	\end{subfigure}

	\caption{
		\textbf{Classification via dimensionality reduction: t-SNE (left) \& UMAP (right).}
		Sets of actions associated with visual observations from 132 positions and orientations in the environment
		for the trained agents of each evolutionary run were passed through the two dimensionality reduction procedures.
		Results are plotted according to:
		1st row: classification labels from the metric-based procedure as illustrated in Fig \ref{fig:classes},
		2nd row: decorrelation time,
		3rd row: directional entropy
		4th row: Gaussian mixture models were fit to the results, redrawing similar class boundaries.
		}
	\label{fig:classes_dimred}
  \end{figure}

\end{document}